\documentclass{article}
\usepackage{arxiv}
\usepackage[utf8]{inputenc} 
\usepackage{hyperref}       
\usepackage{url}            
\usepackage{booktabs}       
\usepackage{amsfonts}       
\usepackage{nicefrac}       
\usepackage{microtype}      
\usepackage{graphicx}
\usepackage{natbib}
\usepackage{doi}
\usepackage{comment}
\usepackage{amssymb} 
\usepackage{color}
\usepackage{wrapfig}
\usepackage{mathtools} 
\usepackage{tikz} 
\usepackage[ruled,vlined]{algorithm2e}
\usepackage{amssymb}

\def\b{\mathbf{b}}
\def\x{\mathbf{x}}

\def\lo{\mathbf{lo}}
\def\hi{\mathbf{hi}}
\def\SDI{{\sc SBox}}
\def\SDICI{{\sc SBoxCI}}
\def\SDT{{\sc SQBox}}
\def\SDTCI{{\sc SQBoxCI}}
\def\CTE{{\sc CTE}}
\def\CTECI{{\sc CTECI}}
\def\sigmahat{\hat{\mathbf{\sigma}}}
\def\Tau{\mbox{T}}
\newtheorem{theorem}{Theorem}
\newtheorem{proposition}{Proposition}
\newtheorem{definition}{Definition}

\begin{document}

\title{Conformal Prediction Intervals for\\Markov Decision Process Trajectories}

\author{Thomas G. Dietterich\\
       School of Electrical Engineering and Computer Science\\
       Oregon State University\\
       Corvallis, OR 97331-5501, USA\\
       {\tt tgd@cs.orst.edu}\\
       \AND
       Jesse Hostetler\\
       SRI International\\
        201 Washington Road
        Princeton, NJ 08540-6449, USA\\
        {\tt jesse.hostetler@sri.com}}
\renewcommand{\shorttitle}{Conformal Prediction Intervals for MDPs}
\maketitle

\begin{abstract}
Before delegating a task to an autonomous system, a human operator may want a guarantee about the behavior of the system. This paper extends previous work on conformal prediction for functional data and conformalized quantile regression to provide conformal prediction intervals over the future behavior of an autonomous system executing a fixed control policy on a Markov Decision Process (MDP). The prediction intervals are constructed by applying conformal corrections to prediction intervals computed by quantile regression. The resulting intervals guarantee that with probability $1-\delta$ the observed trajectory will lie inside the prediction interval, where the probability is computed with respect to the starting state distribution and the stochasticity of the MDP. The method is illustrated on MDPs for invasive species management and StarCraft2 battles. 
\end{abstract}


\section{Introduction}
Suppose you wish to dispatch your personal robot to walk across campus to the cafeteria to pick up your breakfast and bring it back to your room. The robot has learned a policy $\pi$ via reinforcement learning that it will execute to perform this task. However, the probability of running into trouble along the way and failing to achieve the task depends on many factors including the weather, battery charge, time of day, obstacles encountered along the way, and so on. Before you press $GO$, you would like to have confidence that your robot will succeed on the task. 

Let us formalize the definition of ``success'' in terms of the entire state-action-reward trajectory of the robot. Let $\b = (b_1,\ldots, b_H)$ be a vector of length $H$ that describes the performance of the robot along the $H$-step trajectory. For example, $b_t$ could be the immediate reward at time $t$, and a large negative value could indicate that the robot has encountered trouble. A value of $b_H = +1$ in the final time step of the trajectory could indicate successful breakfast delivery. Given the current state $s_0$ of the robot and the world, we would like the robot to provide a prediction interval $\lo \leq \b \leq \hi$ such that with probability $1 - \delta$, the actual vector $\b$ will lie between a lower bound vector $\lo$ and an upper bound vector $\hi$. That is, for all time steps $t$, $lo_t \leq b_t \leq hi_t$. In our example, if $lo_t \geq 0$ for all time steps $t$ and $lo_H \geq 1$, then we know with probability $1-\delta$ the robot will not encounter any trouble and will successfully deliver our breakfast.

In this paper, we present a method for constructing trajectory-wise prediction intervals of this kind. Our method collects a training set of sample trajectories from the (robot) agent executing $\pi$. These trajectories are generated by sampling the starting state from the initial state distribution $P_0$ and then executing $\pi$ for $H$ steps. Along each trajectory $i$, we collect the behavior vector $\b_i$. From these vectors, our method constructs a prediction interval that is a function of the starting state and guarantees finite sample coverage $1 - \delta$ for new trajectories drawn from the same distribution. We prove the correctness of the method and measure its empirical coverage for several values of $\delta$ in two example MDPs---one based on a problem of managing invasive species and the other on managing battles in the video game StarCraft 2. 

We begin by addressing a more fundamental problem, which is of independent interest. Suppose we are given $n$ vectors $\x_1,\ldots,\x_n$ drawn independently from a probability distribution $P$ with support in $\Re^d$. We introduce an algorithm, \SDI, that computes a tight interval $[\lo,\hi]$ such that with probability $1 - \delta$, a new point $\x_{n+1}\sim P(\cdot)$ lies inside the interval $\lo \leq \x_{n+1} \leq \hi$ for $\lo, \hi\in \Re^d$. We prove the correctness of this method. The proof is based on the method of split-conformal prediction \citep{Vovk2005,Shafer2008,Papadopoulos2002}. Our method is closely related to the functional projection framework introduced by \cite{Lei2013}, and it is similar to the method of \cite{Diquigiovanni2021,Diquigiovanni2022}, which was discovered simultaneously. 

Our method treats the prediction problem jointly, whereas previous methods for constructing $n$-dimensional prediction intervals decompose the problem into $n$ one-dimensional problems using the Bonferroni technique. Each one-dimensional interval is constructed to achieve a coverage of $1 - \delta/n$, and this ensures that the concatenation of the $n$ one-dimensional intervals gives joint coverage of $1-\delta$. We will show that the conformal approach gives simultaneous intervals that can be much tighter. 

We then extend this method to generate prediction intervals over the behavior vectors of entire trajectories. This is accomplished in two steps, inspired by \cite{Romano2019}. First, we apply quantile regression to predict the $\delta/2$ and $1 - \delta/2$ quantiles of the behavior vector separately for each time step $t = 1, \ldots, H$ as a function of the starting state $s_0$. Then we apply our second algorithm, \SDT, to adjust these per-time-step predicted quantiles to provide trajectory-wise guarantees. The resulting prediction bounds are ``semi-conditional'' in the sense that the quantile regression bounds are conditioned on the specific starting state $s_0$, but the $1-\delta$ guarantee only holds jointly over all possible starting states sampled according to $P_0$ and therefore might not hold at all on the specific $s_0$.

We introduce two additional heuristic strategies for improving the utility of our prediction intervals. Our first strategy seeks to overcome the semi-conditional nature of the formal guarantee. The guarantee does not rule out the possibility that there is a region $S_{bad}$ of starting states such than when $s_0 \in S_{bad}$, the $[\lo,\hi]$ interval is always violated. Specifically, if the probability of these failure cases $P_0(s_0 \in S_{bad}) \leq \delta$ and if all other starting states satisfy the $[\lo,\hi]$ interval, the formal guarantee still holds. We will present a heuristic strategy for checking that the failure cases do not cluster in this way but that instead they are well distributed throughout the state space. If the distribution of failure cases is independent of the starting state, then the behavior of the semi-conditional interval will be much closer to the behavior of an ideal conditional interval.

Our second strategy addresses another shortcoming of the formal guarantee. The formal $1-\delta$ guarantee states that the \emph{expected} coverage of the $[\lo,\hi]$ interval will be $1-\delta$, where the expectation is taken with respect to the randomly-drawn set of training trajectories. But this means that the actual coverage based on a particular training set may be lower or higher. In safety-critical applications, we might want a stronger guarantee. Our second heuristic strategy strengthens the split-conformal method so that coverage guarantee will hold for fraction $1-\delta$ of the randomly-drawn training sets. 

Providing prospective guarantees of AI agent behavior is an important step toward trustworthy artificial intelligence. Most work on trustworthy AI focuses on providing faithful and informative explanations for the observed behavior of an AI system \textit{retrospectively} \citep{Ribeiro2016WhySI,Adadi2018,Puiutta2020,Fuxin2021}. But such hindsight explanations are not useful at the point in time where the human decision maker must decide whether to press $GO$. The methods introduced in this paper address this need by providing a form of prospective explanation along with a useful statistical guarantee of correctness. 

\section{Related Work}

\begin{figure*}
    \centering
    \includegraphics[width=2.6in]{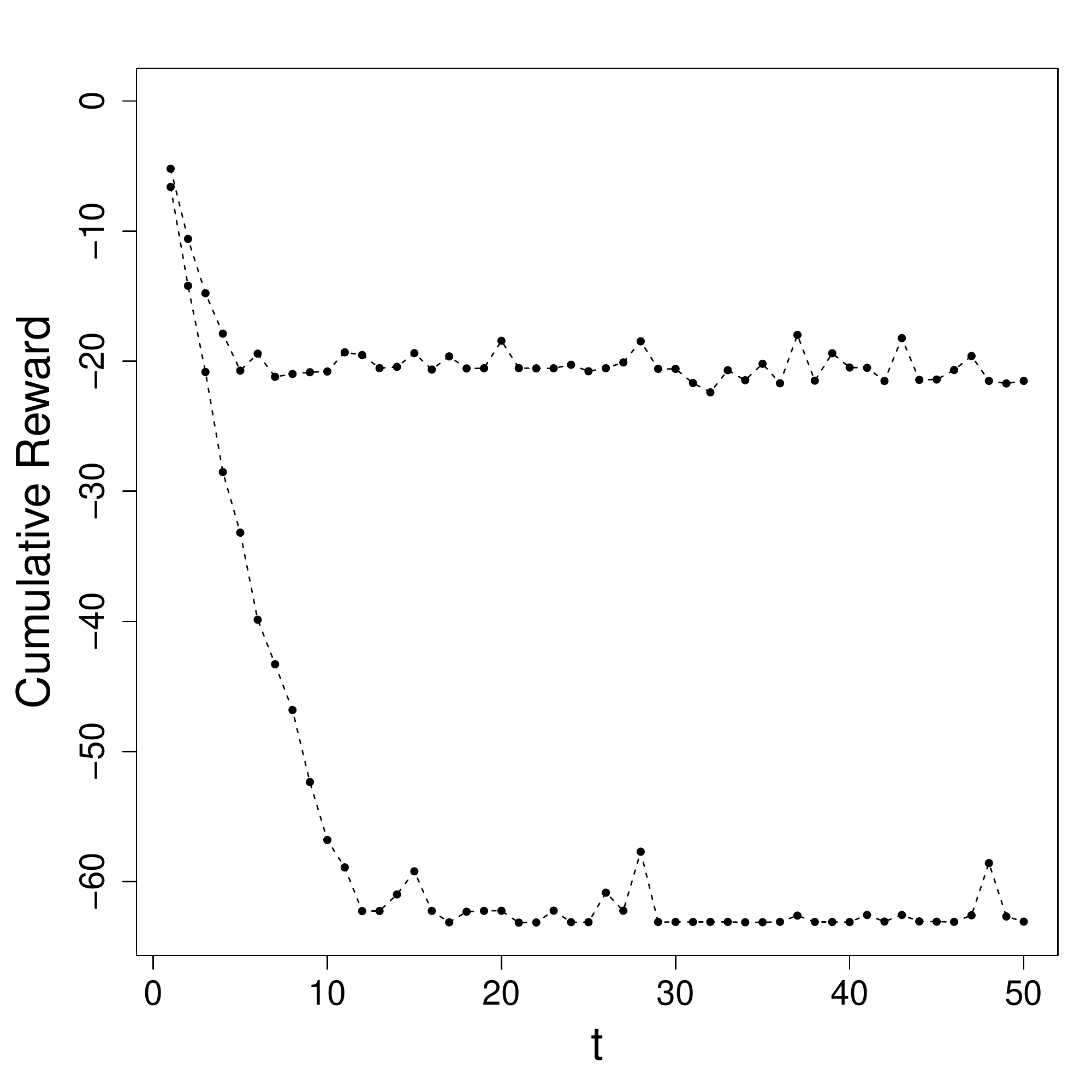} \hspace{0.2in}
    \includegraphics[width=2.6in]{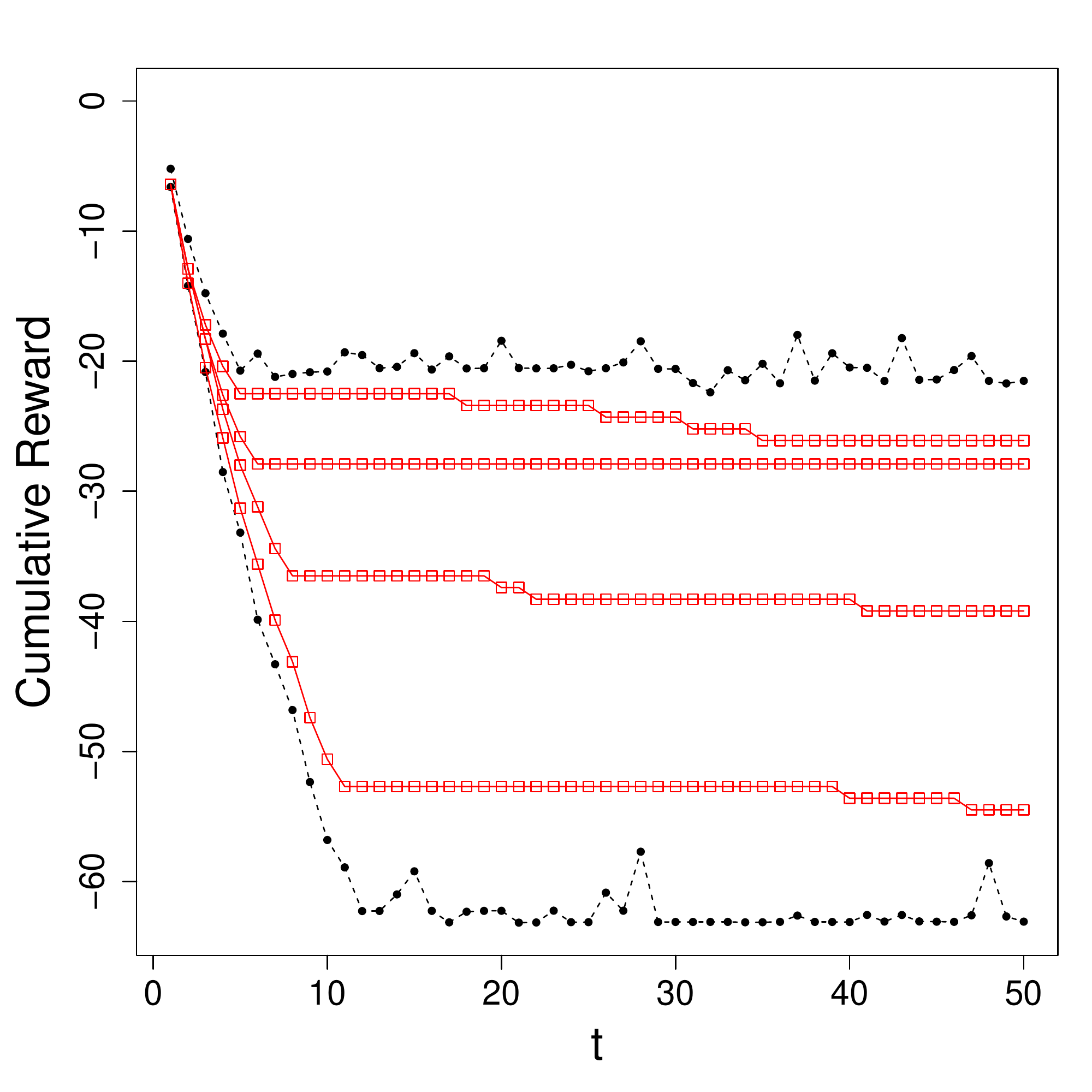}
    \caption{Example trajectory-wise prediction interval for an invasive species management MDP starting in a state with 5 invasive plants and 2 native plants. Left: Black dotted lines show show the 80\% conformalized prediction interval. Right: Red squares show four actual trajectories from this starting state. The values are plotted from $t=1$ to $t=50$ reflecting the results of applying actions at times $t=0$ through $t=49$.}
    \label{fig:trajectory-example}
\end{figure*}

The prediction problem that we address in this paper is related to, but distinct from, several previously-studied problems. Figure~\ref{fig:trajectory-example} shows an example of the kind of prediction interval that our method provides. We are given an MDP and a fixed policy $\pi$ that seeks to control and eliminate an invasive plant species in a river network (see Section~\ref{sec:tamarisk} for details). For this example, we have defined the elements of the behavior vector to be the total reward received so far along the trajectory: $b_t := \sum_{u=1}^t r_u$. The rewards are always negative, and they reflect the total cost of the actions that have been taken from times $t=0$ to $t=49$ to eliminate the invasive species.  The two black lines correspond to the upper and lower prediction bounds for a trajectory starting in state $s_0$ in which there are initially 5 invasive plants and 2 native plants. Our algorithm guarantees that the actual trajectory will lie within these two lines with probability 0.90. The prediction interval depends on the difficulty of the starting state. This particular starting state is quite challenging, so the prediction interval is wide. If the starting state contained no invasive plants, then the interval would be narrow and remain close to zero. A decision maker could use this interval to get a sense of the total time and cost required to control the invasive plant. If these are too large, the decision maker could consider policy alternatives. In this case, the policy $\pi$ is based on a very limited budget in each time step. The decision maker could consider borrowing funds to increase the budget in the early time steps and reduce it in later time steps. The decision maker could also conclude that it is not cost-effective to control the invading plant and decide to spend the money on other conservation priorities. 

We know of no previous work that gives prediction intervals for MDP trajectories. Instead, most work has focused on giving confidence intervals on the value function or the Q function. For example, \cite{Even-dar2003} apply the Hoeffding bound to obtain a finite-sample confidence interval on the value function $V^h(s)$ for being in state $s$ with $h$ steps to go to reach the final horizon $H$. They derive related confidence intervals for $Q^h(s,a)$, the value of being in state $s$ and performing action $a$ with $h$ steps to go. To see how these differ from our prediction intervals, let us define the behavior value $b_t = \sum_{u=t}^H r_u$ to be the sum of future rewards up to horizon $H$ (the ``reward-to-go"). With this definition, the \textit{expected value} of $b_t$ is the fixed-horizon value function $V^{H-t}(s_t)$ for state $s_t$ with $H-t$ steps remaining to the horizon. Applying our method will instead produce a prediction interval that will contain the \textit{actual reward-to-go} with probability $1-\delta$ all along the trajectory based on the initial state $s_0$. In short, our prediction intervals bound the actual behavior of the MDP rather than its expected value. 

A second relevant line of research is distributional reinforcement learning \citep{Morimurat2010,Bellemare2017,Dabney2017,Dabney2018,bdr2022}. Distributional RL methods model the cumulative distribution function of the reward-to-go. By choosing the $\delta/2$ and $1-\delta/2$ quantiles of the CDF, these methods could easily compute a $1-\delta$ prediction interval on the actual reward-to-go. However, these methods only provide approximate quantiles, so they cannot give a guarantee that the actual behavior will fall within the prediction interval. An interesting topic for future research is to find a method, perhaps based on conformal prediction, to calibrate these quantiles to provide finite-sample coverage guarantees. One of the motivations of these authors was to make it easy to optimize risk-sensitive objectives such as the Conditional Value at Risk (CVaR). Their experiments show that they are improving the CVaR, but again because the CDF quantiles are approximate, they cannot provide finite-sample guarantees on the CVaR either.

One method that could be extended to provide predictions along a trajectory is the technique of \textit{generalized value functions} \citep{Sutton2011}. A GVF is a quantity, similar to a value function, that is predicted at each state in an MDP and updated recursively by generalizing the Bellman equation. It would be straightforward to define GVFs for any desired behavior function defined over the next $h$ time steps, for $h=1, \ldots, H$ \citep{Danesh2021}. The methods of distributional reinforcement learning could then be applied to estimate the CDFs of these GVFs and produce approximate prediction intervals. However, as with the distributional RL discussed above, these would not provide coverage guarantees.

\section{Trajectory-Wise MDP Prediction Intervals}

We present our method for trajectory-wise prediction intervals in four steps. First, we review the basic method of conformal prediction. Then we introduce our first method \SDI{} that provides multivariate prediction intervals. In the third step, we shift to the regression setting and review the method of conformalized quantile regression developed by \cite{Romano2019}. Finally, we combine \SDI{} and conformalized quantile regression to create our second method, \SDT{}, which produces trajectory-wise prediction intervals for MDP policies. 

\subsection{Conformal Prediction}
Suppose there are $n+1$ points in $\Re$, $x_1, \ldots, x_n, x_{n+1}$, drawn iid from some distribution $P$. Given the first $n$ points, we wish to output an upper bound $hi$ such that with probability $1-\delta$, $x_{n+1} \leq hi$.  Conformal prediction begins by sorting the first $n$ points into ascending order. Let $x_{(i)}$ denote the $i$th value in ascending order (also known as the $i$th order statistic). 
\begin{proposition} \label{prop:cp}
Let $hi := x_{(\lceil (1-\delta)(n+1)\rceil)}$. If $1/(n+1) \leq \delta < 1$, then with probability $1-\delta$, $x_{n+1} \leq hi$.
\end{proposition}
Proof: The full proof follows Lemma 2 (Inflation of Quantiles) of \cite{Romano2019}. We give the intuition here. Suppose we sorted all $n+1$ points to obtain the order statistics $x'_{(1)}, \ldots, x'_{(n+1)}$. Let $x^* := x'_{\lceil(1-\delta)(n+1)\rceil}$, which is the first value greater than or equal to the $1-\delta$ quantile of the $x'$ points. Let $r'$ be the rank (position) of $x_{n+1}$ in the $n+1$ order statistics. Because $x_1, \ldots, x_{n+1}$ are exchangeable, $r'$ is uniformly distributed on the integers $1, 2, \ldots, n, n+1$.  By the definition of a quantile, the probability that $x_{n+1} \leq x^*$ is at least $1 - \delta$. What quantile must $x^*$ have with respect to the original order statistics $x_{(1)}, \ldots, x_{(n)}$? There are only $n$ elements in this list, so we want the $(1-\delta)\frac{n+1}{n}$ quantile, which is the same index $hi = x_{(\lceil (1-\delta)(n+1)\rceil)}$. The value of $\delta$ must be $\geq 1/(n+1)$ to ensure that $\lceil(1 - \delta)(n+1)\rceil \leq n$. $\square$

Conformal prediction was introduced by \cite{Vovk2005} and \cite{Shafer2008}, although the observation that $r'$ is uniformly distributed among the order statistics was made earlier by \cite{Geisser1993}. The beauty of  Proposition~\ref{prop:cp} is that it gives a finite sample guarantee that applies to any distribution.

\subsection{Multivariate Prediction Intervals}

Now let's consider the multidimensional case. Suppose we are given $n$ vectors $\x_1,\ldots,\x_n$ of dimension $d$ drawn independently from a probability distribution $P$ with support in $\Re^d$. We will denote the $j$-th element of $\x_i$ by $x_{i,j}$. Given a confidence parameter $\delta$, we wish to produce a prediction interval $[\lo,\hi]$ such that with probability at least $1-\delta$, a new vector $\x_{n+1}$ also drawn from $P$ will fall inside the interval: $\lo \leq \x_{n+1} \leq \hi$. We want this multivariate prediction interval to be small in the sense that the probability that $\x_{n+1} \in [\lo,\hi]$ should not be substantially larger than $1-\delta$. 

To apply conformal prediction, we need to reduce this $d$-dimensional problem to a 1-dimensional quantity. We achieve this as follows. We start by using the first $m$ vectors to estimate the mean $\hat{\mu}_j$ and sample standard deviation $\hat{\sigma}_j$ along each dimension $j=1,\ldots, d$. The final prediction interval will be determined by a parameter $\beta$ such that the width of the prediction interval for dimension $j$ is $\beta$ scaled by the sample standard deviation for dimension $j$: $[\hat{\mu}_j - \beta \hat{\sigma}_j, \hat{\mu}_j + \beta \hat{\sigma}_j]$. The value of $\beta$ will be selected via conformal prediction using the remaining $n-m$ data points. Let $x'_{i,j} = (x_{i,j} - \hat{\mu}_{i,j})/\hat{\sigma}_j$ for $i=m+1, \ldots, n$ and $j=1, \ldots, d$. (If $\hat{\sigma}_j = 0$, we set it to the smallest of the nonzero $\hat{\sigma}_j$ values. If all $\hat{\sigma}_j$ values are zero, we abort.) This standardizes each dimension $j$ separately. Now we define $c_i = \max_j |x'_{i,j}|$ to be the maximum distance (in standard units) of $x'_{i,j}$ away from the origin. We will refer to such quantities as ``exceedances'' for reasons that will become clear in later sections. By computing the order statistics $c_{(i)}$ and selecting $\beta := c_{(\lceil (1-\delta)(n-m+1)\rceil)}$, we obtain the $d$-dimensional prediction interval $[\lo,\hi]$. We call this procedure the Scaled Box or \SDI{}. The pseudo-code is shown in Algorithm~\ref{alg:SDI}. 

 \begin{algorithm}
\DontPrintSemicolon
\KwData{$\x_1, \ldots, \x_n$: data points in $\Re^d$\\
        $\delta \geq 1/(n-m+1)$: error probability\\
        $m$: sample size for estimating means and variances}
\KwResult{$\lo,\hi \in \Re^d$: prediction interval limits}
\Begin{
  Use the first $m$ points to estimate the mean and standard deviation along each dimension $j$:\;
  $\hat{\mu} \leftarrow \frac{1}{m} \sum_{i=1}^m \x_i$ \;
  $\hat{\sigma}_j \leftarrow \sqrt{\frac{1}{m-1} \sum_{i=1}^m (x_{ij} - \hat{\mu}_j)^2}\quad \forall j\in\{1,\ldots,d\}$\;
      Replace zero standard deviations with the smallest nonzero standard deviation:\;
      \lIf{$\hat{\sigma}_j = 0$}{$\hat{\sigma}_j \leftarrow \min_{\{j':\hat{\sigma}_{j'}>0\}} \hat{\sigma}_{j'}$}
  Standardize the remaining data points:\;
  \For{$i \in m+1,\ldots,n$}{
    $x'_{i,j} \leftarrow (x_j - \hat{\mu}_j) / \hat{\sigma}_j \quad \forall j\in\{1,\ldots,d\}$\;
    }
  \For{$i \in m+1,\ldots,n$}{
    Compute the maximum exceedance of each data point:\;
    $c_i \leftarrow \max_j |x'_{ij}|$
    }
   Sort $c_{m+1},\ldots,c_n$ to obtain the order statistics $c_{(1)},\ldots,c_{(n-m)}$\;
   Determine the $(1 - \delta)((n-m+1)/(n-m))$ quantile of the $c$ values:\;
  $\beta \leftarrow c_{(\lceil(1 - \delta)(n-m+1)\rceil)}$\;
  $\lo \leftarrow \hat{\mu} -\beta \hat{\sigma}$\;
  $\hi \leftarrow \hat{\mu} +\beta \hat{\sigma}$\;
  \Return{$[\lo,\hi]$}
}
\caption{\SDI{} (Scaled Box)}
\label{alg:SDI}
\end{algorithm}

\begin{proposition}
The exceedance values $c_{m+1},\ldots,c_n$ are exchangeable.
\end{proposition}
Proof: Because we use the first $m$ data points to compute $\hat{\mu}$ and $\hat{\sigma}$, these are fixed when determining the $c_i$ values. The data points $\x_{m+1}, \ldots, \x_n$ are iid, and the $c_i$ values are invariant to shuffling the order of $\x_{m+1}, \ldots, \x_n$. Hence, they are exchangeable. $\square$

\begin{theorem}\label{th:SDSI}
Let $\x_1, \ldots, \x_n, \x_{n+1}\in \Re^d$ be independent random variables with distribution $P$. Let $[\lo,\hi]$ be the multidimensional interval computed by \SDI{} when applied to $\x_1, \ldots, \x_n$ with $2 \leq m < n$ and confidence parameter $\delta \in [1/(n-m+1),1)$. Then with probability $1-\delta$, $\lo \leq \x_{n+1} \leq \hi$. 
\end{theorem}
Proof: The result follows the same argument as the proof of Proposition~\ref{prop:cp}. $\square$

Note that in this result, we are making heavy use of the distribution-free guarantee from Proposition~\ref{prop:cp}. Because the $c_i$ values are computed as a maximum of $2d$ quantities, they will follow an extreme value distribution, so standard prediction intervals based on Gaussian distributions would not give good results. 

The purpose of scaling according to the standard deviation is to ensure that the prediction interval reflects the ``shape'' of the data. Along some dimensions, the $\x_i$ vectors may vary substantially, and we will want the prediction interval to be wide. Along other dimensions, the $\x_i$ vectors may show little or no variation, in which case we want a tight prediction interval.  

Algorithm~\ref{alg:SDI} is closely-related to the conformal prediction intervals for functional data introduced by \cite{Lei2013}. They consider the problem where the data are generated (conceptually) by first drawing a continuous function $f_i: [0,1] \mapsto \Re$ iid from an unknown distribution $P$ and then projecting that function onto $d$ orthogonal basis functions to obtain $\x_i = (x_{i,1}, \ldots, x_{i,d})$. To define a nonconformity measure over the $\x_i$ vectors, they instantiate the density estimation theory of \cite{Lei2013a} by fitting a mixture of multivariate Gaussian distributions. The nonconformity score $c_i$ for $\x_i$ is then defined as the value of the density estimator evaluated at $\x_i$. Under very general conditions, this gives a statistically efficient conformal prediction interval.

Our \SDI{} algorithm can be viewed as instantiating their Algorithm 2, where the basis functions are the coordinate axes of $\Re^d$, and the nonconformity measure is
\[c(x) = \max_j \frac{|x_j -\hat{\mu}_j|}{\hat{\sigma}_j}.\]
We can view each $|x_j -\hat{\mu}_j| / \hat{\sigma}_j$ as a measure of the nonconformity of $x_{ij}$. Instead of incurring the cost and complexity of fitting a density estimator to these nonconformity measures in $d$-dimensional space, our approach simply takes the maximum of the per-dimension nonconformity measures and applies the standard comformal prediction procedure. 

In a preprint \citep{Diquigiovanni2021} and a paper in press \citep{Diquigiovanni2022}, Diquigiovanni et al.~conduct a thorough study of conformal prediction intervals for functional data. They investigate several methods for rescaling the $x_j$ values including the sample standard deviation and the inter-quartile range. In particular, they introduce the following interesting method. First, they compute the conformalized box interval $[\lo^u,\hi^u]$ using unscaled values $c(x) = \max_j |x_j -\hat{\mu}_j|$ computed on the first $m$ points. They then use those same $m$ points to define $\sigma(j)$ as $(hi^u_j - lo^u_j)/z$, where $z = \sum_j hi^u_j - lo^u_j$ normalizes the $\sigma(j)$ values so that they sum to 1. These $\sigma(j)$ values are then applied to rescale $(x_j - \hat{\mu}_j)$ in the same way that \SDI{} employs $\hat{\sigma}_j$. They replace zero values of $\sigma(j)$ with a small constant prior to normalization. An interesting direction for future work would be to test their methods on MDP trajectories. 
 
\subsubsection{Improving \SDI{} Coverage with Quantile Confidence Intervals}
Initial experiments with \SDI{} (see Appendix~\ref{ap1}) showed that---consistent with Proposition~\ref{prop:cp}---the expected fraction of future vectors covered by the prediction intervals is almost exactly equal to $1-\delta$. However, there is also substantial variation around this mean coverage (see Figure~\ref{fig:t-success}. This naturally arises because the conformal method is estimating a quantile, and that estimate has nonzero variance. Consequently, in a substantial fraction of the cases, the coverage is below $1-\delta$ (and in a similar fraction of cases, the coverage is above $1-\delta$). In safety-critical applications, we want a prediction interval that achieves coverage of at least $1-\delta$ in fraction $1-\delta$ of future trials, where each trial applies the method to a fresh data set. 

To obtain this, we can make a slight modification to \SDI{} to replace the quantile computation 
\begin{quote}
    $\beta \leftarrow c_{(\lceil(1 - \delta)(n-m+1)\rceil)}$
\end{quote}
with
\begin{quote}
    $\beta \leftarrow 1-\delta \mbox{ upper confidence bound on the } (1-\delta)\frac{n-m+1}{n-m}\mbox{ quantile of }c_1, \ldots, c_{n-m}$
\end{quote}
We compute this upper confidence bound using the method of \cite{Nyblom1992}. We will refer to the modified method as \SDICI{}.

\subsection{Conformalized Quantile Regression}
Let us now consider data of the form $(\x_1,y_1),\ldots,(\x_n,y_n), (\x_{n+1},y_{n+1})$ drawn iid from a distribution $P$ with support on $\Re^d\times \Re$. The $\x_i$ are covariates (feature vectors), and the $y_i$ values are the responses. For each value of $\x$, the response $y$ is distributed according to $P(y|\x)$.  We seek a conditional prediction interval $[lo(\x_{n+1}),hi(\x_{n+1})]$ such that with probability $1-\delta$, $lo(\x_{n+1})\leq y_{n+1} \leq hi(\x_{n+1})$. We could fit a least-squares regression function to the data and then apply standard conformal prediction to the residuals. However, \cite{Romano2019} introduced a beautiful method based on first performing quantile regression and then computing a conformal correction to the quantile regression ``residuals''.

Quantile regression methods \citep{Koenker2005,Meinshausen2006} fit functions of the form $\hat{y} = q_\alpha(\x)$ such that $\hat{y}$ is approximately the $\alpha$ quantile of $P(y|\x)$. Using these quantile functions, we can obtain a conditional prediction interval by setting $lo(\x)=q_{\delta/2}(\x)$ and $hi(\x)=q_{1 - \delta/2}(\x)$. However, such an interval does not provide a finite-sample guarantee. To obtain a guarantee, \cite{Romano2019} use the first $m$ data points to fit the quantile functions $q_{\delta/2}$ and $q_{1-\delta/2}$. Then for the remaining points, they compute the exceedances $c_i = \max\{q_{\delta/2}(\x_i) - y_i,\; y_i - q_{1 - \delta/2}\}$. Finally, they compute the order statistics $c_{(1)}, \ldots, c_{(n-m)}$ and let $\hat{c} := c_{(\lceil (1 - \delta)(n - m + 1)\rceil)}$ be the conformal adjustment. The prediction interval for $y_{n+1}$ is then computed as $[q_{\delta/2}(\x_{n+1})-\hat{c}, q_{1 - \delta/2}(\x_{n+1}) + \hat{c}]$. 

\begin{theorem}[\cite{Romano2019}]\label{th:romano}
If $(\x_i,y_i)$ for $i=1,\ldots,n+1$ are exchangeable, then the probability that
\[q_{\delta/2}(\x_{n+1})-\hat{c}\leq y_{n+1} \leq q_{1 - \delta/2}(\x_{n+1}) + \hat{c}\]
is at least $1-\delta$. 
\end{theorem}
A nice aspect of this technique is that if the predicted quantiles are too wide, the $c_i$ values of will be negative, and the conformal correction $\hat{c}$ will also be negative. Hence, the conformal interval will be tighter than the interval produced by quantile regression. In fact, \cite{Romano2019} found that using a slightly larger value $\delta' > \delta$ for the quantile regressions can result in slightly tighter conformal prediction intervals. 

As we discussed in the introduction, it is important to note that the guarantee is with respect to a random pair $(\x_{n+1},y_{n+1})$ drawn from the joint distribution $P(\x_{n+1},y_{n+1})$. This is a marginal guarantee rather than a conditional guarantee. A conditional guarantee would hold for $y_{n+1}$ drawn according to $P(y_{n+1}|\x_{n+1})$. Unfortunately, \cite{Vovk2012} and \cite{Barber2019} show that conditional guarantees are impossible to provide in general. 
We call the conformal quantile interval a \textit{semi-conditional interval}, because the quantile functions $q_{\delta/2}(\x_{n+1})$ and $q_{1 - \delta/2}(\x_{n+1})$ are conditioned on $\x_{n+1}$, but the conformal correction $\hat{c}$ is not. 

\subsection{Prospective Trajectory-wise Prediction Intervals for MDPs}

We now combine the \SDI{} algorithm with conformal quantile regression to construct prospective performance guarantees for the behavior of a policy executed on an MDP. By \textit{prospective}, we mean that the future performance is predicted based only on the starting state $s_0$. This guarantee can then be used by the human decision maker to decide whether to let the AI system execute the policy $\pi$ autonomously for $H$ steps. In our experiments, we define the \textit{behavior} as the cumulative $h$-step reward, but we stress that our method can be applied to any quantity that can be measured at each time step along a trajectory, including state variables, action statistics, and generalized value functions \citep{Sutton2011}. 

\begin{algorithm}[ht]
\DontPrintSemicolon
\KwData{$\tau_1, \ldots, \tau_{n+1}$: sampled trajectories\\
        $\b_1, \ldots, \b_{n+1}$: corresponding behavior vectors\\
        $l$: subsample size for fitting the quantile regressions\\
        $m$: subsample size for estimating the standard deviations\\
        $\delta \geq 1/(n-m-l+1)$: desired error probability\\
        $\delta' > \delta$: target error probability for quantile regressions
}
\KwResult{$\lo,\hi \in \Re^H$: prediction interval for $\b_{n+1}(\tau_{n+1})$}
\Begin{
  \For{$t \in 0, \ldots, H-1$}{
    Define the quantile regression training set $D_t \leftarrow \{(S_0(\tau_1), b_{1,t}), \ldots, (S_0(\tau_l), b_{l,t})\}$\;
    Fit $q_{lo,t}$ to $D_t$ to estimate the $\delta'/2$ quantile of $\b_{\cdot,t}$\;
    Fit $q_{hi,t}$ to $D_t$ to estimate the $1 - \delta'/2$ quantile of $\b_{\cdot,t}$\;
    Compute exceedance trajectories $\x_i$:\;
    \For{$i \in l+1,\ldots,n$}{
    Let $x_{i,t} \leftarrow \max\{0, \; q_{lo,t}(S_0(\tau_i)) - b_{i,t},\; b_{i,t} - q_{hi,t}(S_0(\tau_i))\}$}
    }
  Use $\b_{l+1}, \ldots, \b_{l+m}$ to estimate the standard deviation at each time step $t$:\;
  $\hat{\sigma}_t \leftarrow \sqrt{\frac{1}{m} \sum_{i=l+1}^{l+m} x_{i,t}^2}\quad \forall t\in\{1,\ldots,H\}$\;\vspace*{0.05in}
  Replace zero standard deviations with the smallest nonzero standard deviation:\;
  \lIf{$\hat{\sigma}_t = 0$}{$\hat{\sigma}_t \leftarrow \min_{\{t':\hat{\sigma}_{t'}>0\}} \hat{\sigma}_{t'} \quad \forall t$}
  Standardize the remaining exceedance vectors:\;
  \For{$i \in l+m+1,\ldots,n$}{
    $x'_{i,t} \leftarrow x_{i,t} / \sigmahat_t \quad \forall t \in 0, \ldots, H -1$\;
    }
  Compute the maximum exceedance of each trajectory:\;
  \For{$i \in l+m+1,\ldots,n$}{
      $c_i \leftarrow \max_t x'_{i,t}$
    }
   Sort $c_{l+m+1},\ldots,c_n$ to obtain the order statistics $c_{(1)},\ldots,c_{(n-l-m)}$\;
   Determine the $(1 - \delta)((n-l-m+1)/(n-l-m))$ quantile of the $c$ values:\;
  $\beta \leftarrow c_{(\lceil(1 - \delta)(n-l-m+1)\rceil)}$\;
  \For{$t \in 0,\ldots, H-1$}{
    $lo_t \leftarrow q_{lo,t}(S_0(\tau_{n+1})) - \beta \hat{\sigma}_t$\;
    $hi_t \leftarrow q_{hi,t}(S_0(\tau_{n+1})) + \beta \hat{\sigma}_t$
    }
  \Return{$[\lo,\hi]$}
}
\caption{\SDT{} (Scaled Quantile Box)}
\label{alg:SDSI-trajectory}
\end{algorithm}

\begin{figure*}
    \centering
    \includegraphics[width=2.9in]{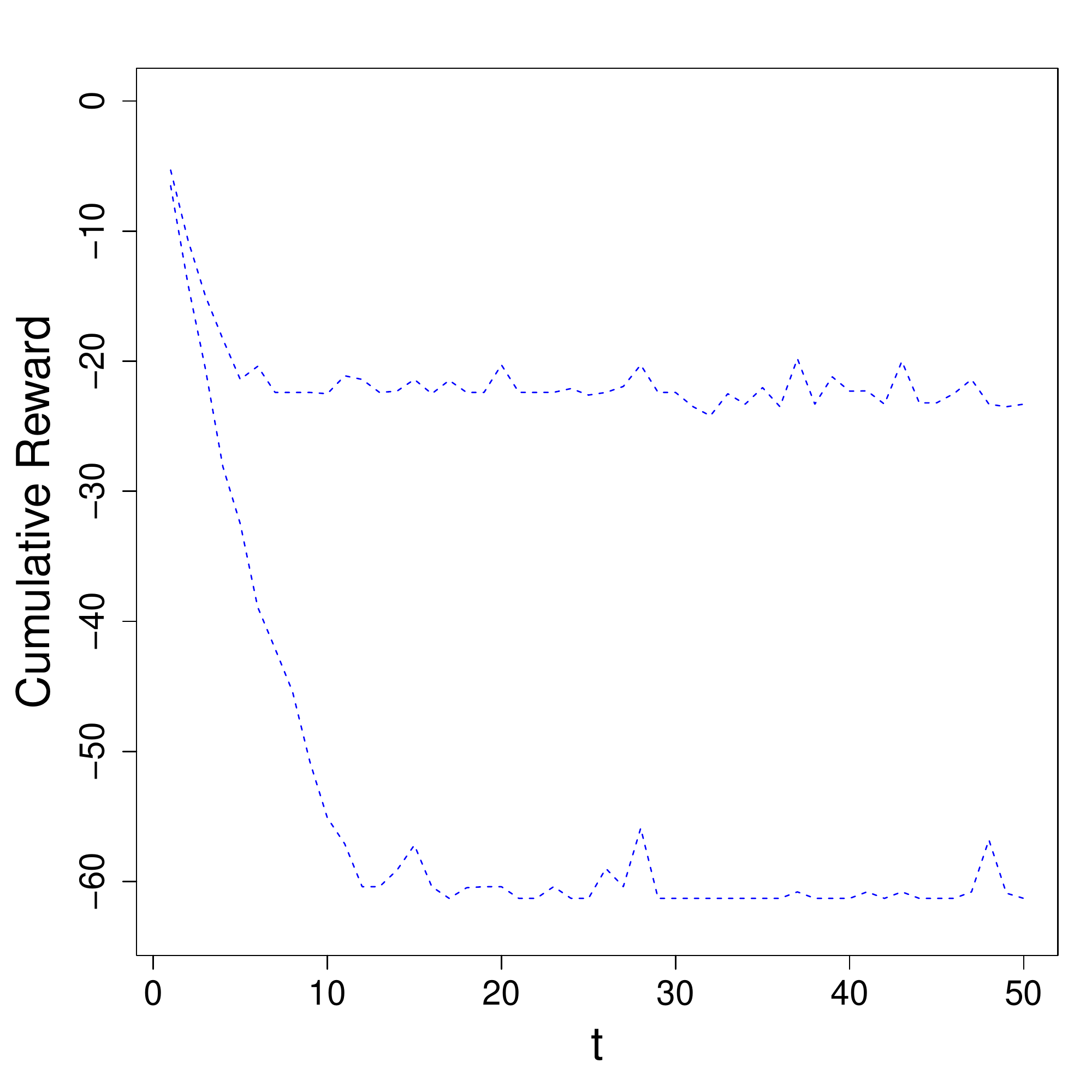} 
    \includegraphics[width=2.9in]{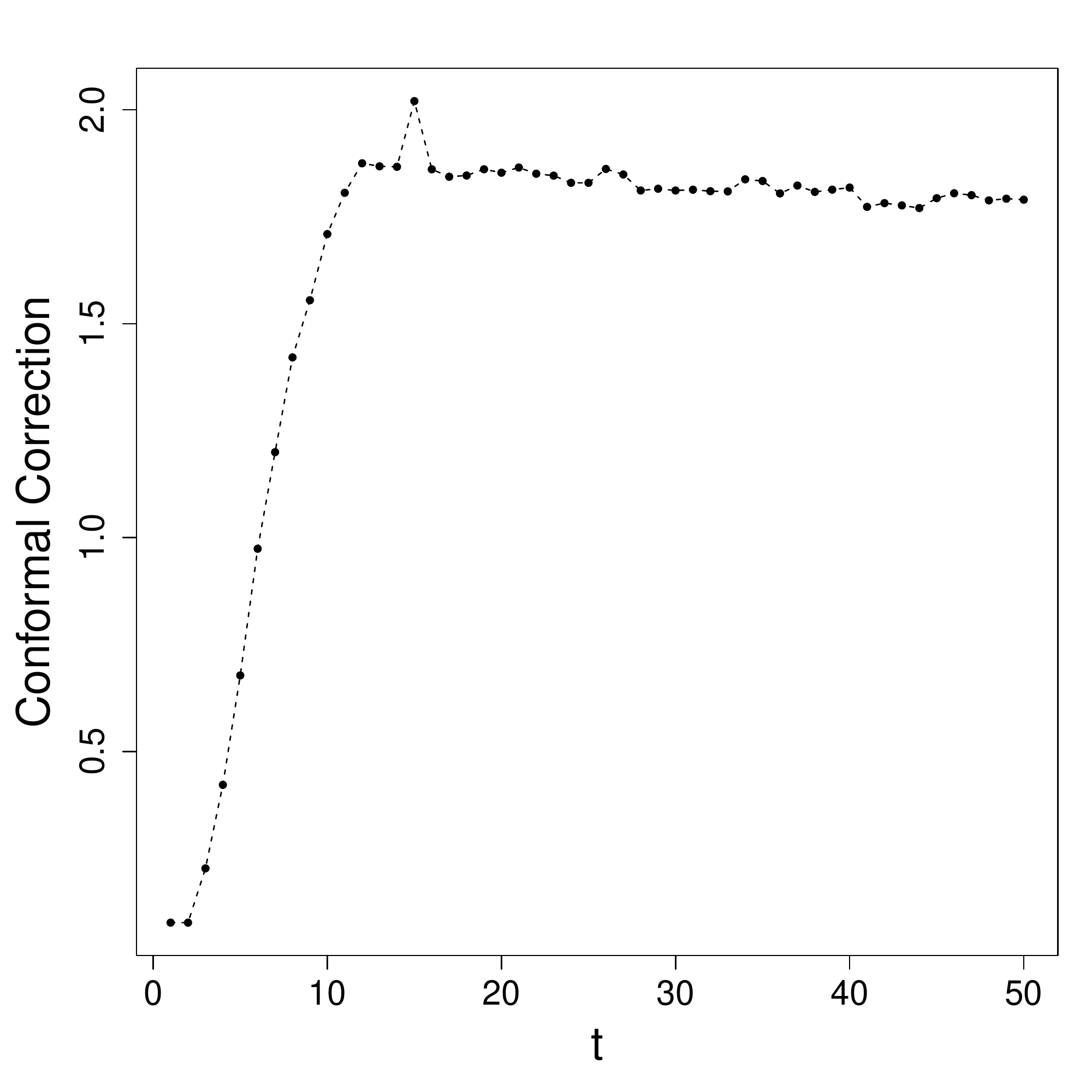}
    \includegraphics[width=2.9in]{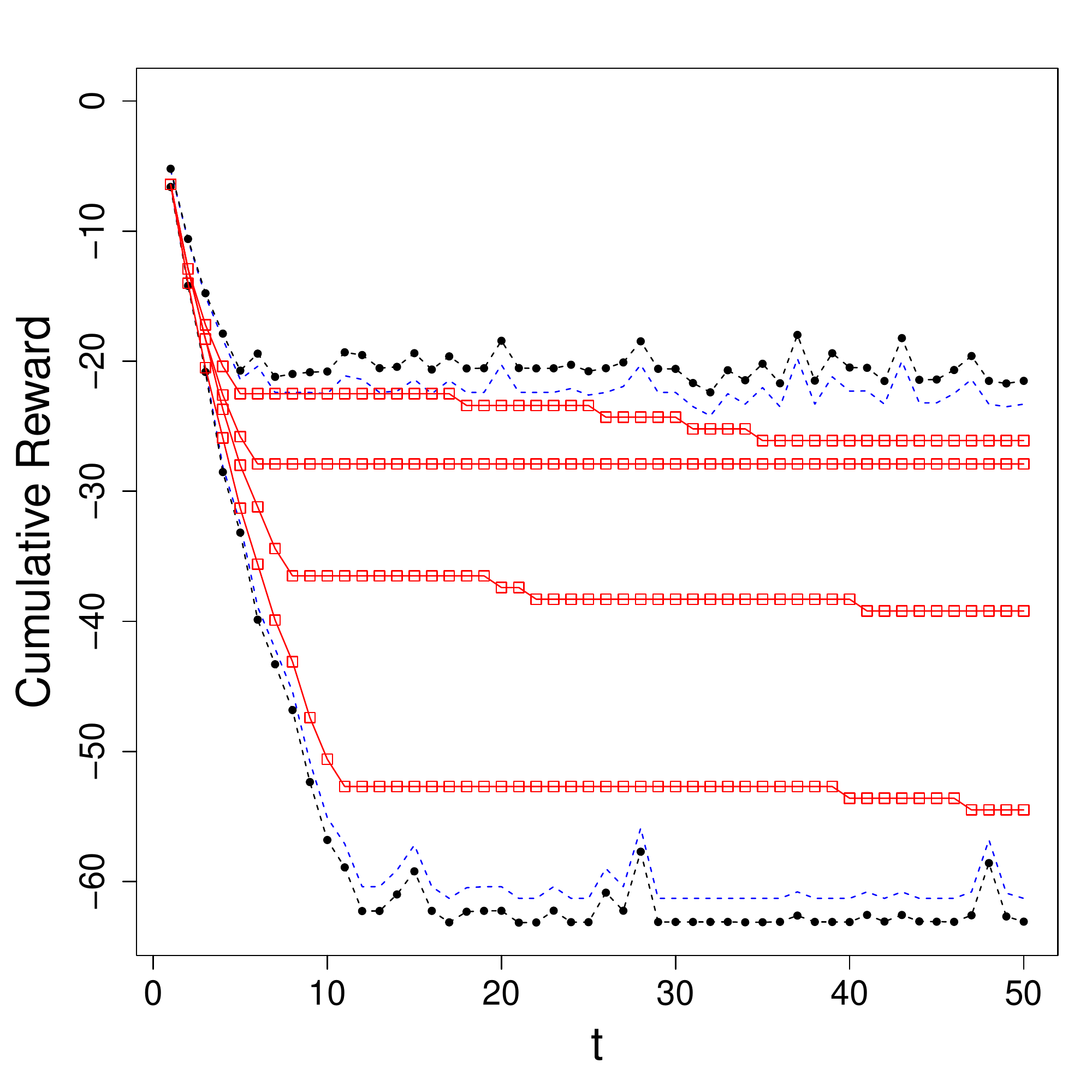}
    \caption{Illustration of the conformal adjustment process for a given starting state $s_0$. Top left: Initial quantile regressions for $s_0$ (blue); Top right: Conformal corrections $\beta\sigmahat$ (black); Bottom: initial quantile regressions (blue), final trajectory-wise predictions (black), and four actual trajectories (red) all starting in $s_0$.}
    \label{fig:conformal-adjustment-example}
\end{figure*}

Suppose we have an agent interacting with an environment and we model this interaction as a discrete-time Markov Decision Process with state space $S$ and starting state distribution $P_0(\cdot)$. Let $\pi$ be the policy that is executed by the agent. Let $\Tau$ be the space of possible trajectories, and let $\tau \in \Tau$ be a random variable that represents a trajectory generated by first selecting $s_0 \sim P_0(\cdot)$ and then executing $\pi$ for $H$ steps. Each $\tau$ is a sequence of state, action, and reward triples plus the final state: $(\langle s_0, a_0, r_0\rangle, \ldots, \langle s_{H-1}, a_{H-1}, r_{H-1}\rangle, s_H)$. Define the function $S_0(\tau)$ to return the starting state $s_0$ of $\tau$.

We now define a \textit{behavior function} $b_t(\tau): \Tau \mapsto \Re$ that maps from the space of trajectories $\Tau$ to a real-valued quantity that summarizes some interesting aspect of the behavior of the agent at time $t$. The value of $b_t(\tau)$ can depend on all or part of the trajectory $\tau$. Let $\b(\tau) = (b_1(\tau), \ldots, b_H(\tau))$ denote the behavior vector computed from $\tau$. We will drop the dependence on $\tau$ when it is clear from context.

Let $\tau_1, \ldots, \tau_n$ be a collection of $n$ trajectories, each generated by sampling a starting state from $P_0$ and then executing $\pi$ for $H$ steps. Let $\b_1, \ldots, \b_n$ be the corresponding behavior vectors. Our goal is to compute a trajectory-wise interval $[\lo,\hi]$ such that if $\tau_{n+1}$ is a new trajectory sampled in the same way, its behavior vector will be contained inside the interval
\[\lo \leq \b_{n+1} \leq \hi\]
with probability $1-\delta$. 

The main idea of our approach is to first define an ``inner'' interval via quantile regression and then apply the ideas from \SDI{} with this inner interval playing the role of $\mu$. We begin by fitting quantile regression functions $q_{lo,t}$ and $q_{hi,t}$ to predict the $\delta'/2$ and $1 - \delta'/2$ quantiles of $b_t$ for $t=1,\ldots,H$. We use only the first $l$ trajectories for this purpose. (We set $\delta'$ to a value somewhat larger than our overall error value $\delta$ for reasons we will explain below.) Then, using the remaining $n-l$ trajectories, we compute the difference between the predicted quantile and the observed value for each behavior trajectory and time step. This exceedance measures the amount by which the observed value of $b_t(\tau)$ falls outside the predicted ``inner'' interval $[q_{lo,t}(S_0(\tau)), q_{hi,t}(S_0(\tau))]$. 

\begin{definition}
The \emph{exceedance} $x_{i,t}$ of behavior vector $\b_i$ for trajectory $\tau_i$ at time $t$ is defined as
\begin{align} \nonumber
x_{i,t} = \max\{&0,\\ \nonumber
&q_{lo,t}(S_0(\tau_i)) - b_{i,t},\\ \nonumber
& b_{i,t} - q_{hi,t}(S_0(\tau_i))\}.
\end{align}
The \emph{exceedance trajectory} $\x_i = (x_{i,1}, \ldots, x_{i,H})$ is the sequence of exceedance values for $\b_i$.
\end{definition}

The next step is to apply \SDI{} to the exceedance trajectories. As in \SDI{}, we analyze the first $m$ exceedance trajectories to compute the standard deviation $\hat{\sigma}_j$ along each dimension $j=1,\ldots, H$. 


We then standardize the exceedance trajectories to obtain standardized exceedance vectors $\x'_{l+m+1}, \ldots, \x'_{n}$, compute $c_i$ as the maximum value of each standardized exceedance vector, compute the order statistics $c_{(l+m+1)}, \ldots, c_{(n)}$, and set $\beta$ equal to $c_{(\lceil (1 - \delta)(n - l - m + 1)\rceil)}$. The conformal prediction interval for time step $t$ is computed as
\begin{align}
    lo_t &\leftarrow q_{lo,t}(S_0(\tau_{n+1})) - \beta \hat{\sigma}_t\\ \nonumber
    hi_t &\leftarrow q_{hi,t}(S_0(\tau_{n+1})) + \beta \hat{\sigma}_t. \nonumber
\end{align}

Algorithm~\ref{alg:SDSI-trajectory} provides the pseudo-code for \SDT{}.

Figure~\ref{fig:conformal-adjustment-example} shows an example of computing the conformal correction. The upper left panel shows the trajectory-wise ``inner'' interval predicted by the quantile regressions for a particular starting state $s_0$. The upper right panel shows the conformal corrections $\beta \sigmahat$. Note that the same correction is applied to the quantile regressions for all starting states; the correction does not depend on $s_0$. The lower panel shows the conformalized trajectory-wise interval along with four actual trajectories from a separate test set (all starting in the same $s_0$). 

\begin{theorem} If the trajectories $\tau_1,\ldots, \tau_{n+1}$ are generated by sampling starting states  $s_0$ iid from $P_0$ and following fixed policy $\pi$ for $H$ steps, then behavior vector $\b_{n+1}(\tau_{n+1})$ will fall within the prediction interval $[\lo,\hi]$ returned by \SDT{} with probability $1 - \delta$.
\end{theorem}
Proof: Because the quantile regressions are computed on a disjoint set of $l$ trajectories and $\{\hat{\sigma}_t\}_1^H$ are estimated from a disjoint set of $m$ trajectories, the remaining $n-l-m$ trajectories and trajectory $\tau_{n+1}$ are exchangeable. Consequently, the behavior vectors $\b_{l+m+1},\ldots, \b_{n}, \b_{n+1}$ are exchangeable and so are their exceedances $\x_{l+m+1}, \ldots, \x_{n}, \x_{n+1}$. Hence, exceedance trajectory $\x_{n+1}$ computed from $\b_{n+1}$ will fall within the scaled values $[-\beta \hat{\sigma},+\beta\hat{\sigma}]$ by applying the conformal argument of Theorem~\ref{th:SDSI}. $\square$

In preliminary experiments, we found that when $\delta' = \delta$, the ``inner'' interval computed by the quantile regressions may be so wide that many of the exceedances $b_{i,t}$ are zero. Consequently, the $\sigmahat_t$ values are poorly-estimated and the quantile corrections $\beta \sigmahat_t$ do not give good results. Therefore, we recommend using a value of $\delta'$ that is large enough that most exceedances are nonzero. 

\subsubsection{Algorithm \SDTCI{}}
As with \SDI{} and \SDICI{}, we can improve the coverage of \SDT{} by replacing the computation of $\beta$ in \SDT{} with a $1-\delta$ upper confidence bound on the desired quantile. Specifically, we replace
\begin{quote}
    $\beta \leftarrow c_{(\lceil(1 - \delta)(n-l-m+1)\rceil)}$
\end{quote}
with
\begin{quote}
    $\beta \leftarrow 1-\delta \mbox{ upper confidence bound on the } (1-\delta)\frac{n-l-m+1}{n-l-m}\mbox{ quantile of }c_1, \ldots, c_{n-l-m}$
\end{quote}
We call \SDT{} with this modification \SDTCI{}. 

\subsection{Conformal Total Exceedance as an Alternative}
Some decision makers might prefer a different representation of the future behavior of policy $\pi$. Rather than providing a multidimensional prediction interval $[\lo,\hi]$ that is guaranteed (with probability $1-\delta$) to contain the future trajectory, we could instead provide a multidimensional interval $[\lo,\hi]$ and a bound on the total amount by which trajectory $\tau_{n+1}$ will exceed that interval. Specifically, for $i=l+1,\ldots, n+1$, let $c_i = \sum_{t=1}^H x_{i,t}$ be the total exceedance of trajectory $\tau_i$, and let $\hat{c}$ denote the $\lceil(1-\delta)(n-l+1)/(n-l)\rceil$ quantile of the total exceedances $c_{l+1},\ldots,c_{n}$. Then we could present to the user the results of the quantile regression
\begin{align}\nonumber
lo_t &= q_{lo,t}(S_0(\tau_{n+1})) \;\; &\forall t \in \{1,\ldots,H\}\\ \nonumber
hi_t &= q_{hi,t}(S_0(\tau_{n+1})) \;\; &\forall t \in \{1,\ldots,H\}
\end{align}
along with $\hat{c}$. As we do not need to estimate $\hat{\mu}$ and $\hat{\sigma}$, we can use all $n-l$ trajectories to determine $\hat{c}$. When fitting the quantile regressions, we recommend fitting $q_{lo,t}$ to predict the $\delta/2$ quantile, and $q_{hi,t}$ to predict then $1 - \delta/2$ quantile, rather than using a value $\delta' >\delta$. This is because we want the total exceedances to be small. We will call this algorithm \CTE{} (Conformalized Total Exceedance).

\begin{proposition}
With probability $1-\delta$, the total exceedance $c_{n+1} \leq \hat{c}$.
\end{proposition}
Proof: Because the exceedance vectors $\x_{l+1},\ldots,\x_{n},\x_{n+1}$ are exchangeable, their total exceedances $c_{l+1}, \ldots, c_{n},c_{n+1}$ are exchangeable. Hence, the standard conformal argument applies. $\square$

As with \SDTCI{}, we can replace the $1 - \delta/2$ quantile with a $1-\delta$ upper confidence bound on the quantile. We will denote this method by \CTECI{} (Conformalized Total Exceedance with upper Confidence Interval).

\section{Experimental Studies}
We conducted two experimental studies. The first study examines the behavior of \SDI{} and \SDICI{} by comparing them to the more traditional approach of computing separate prediction intervals for each dimension and applying the Bonferroni correction. We will show that \SDICI{} gives much tighter prediction intervals while achieving correct prediction interval coverage in cases where there are correlations among the $x_{\cdot{}1}, \ldots, x_{\cdot{}d}$. 

The second study examines the performance of \SDT{} and \SDTCI{} on two sequential decision-making problems: Tamarisk and Starcraft. In both domains, we have implemented a fixed policy and collected thousands of trajectories by sampling according to a starting state distribution $P_0$ and then applying the policy for $H$ time steps; $H=50$ for Tamarisk and $H=57$ for Starcraft. We will see that \SDT{} and \SDTCI{} give excellent performance on the Tamarisk problem and \SDTCI{} gives excellent performance on the Starcraft task. Both are much better than the quantile regression baseline.  We also evaluate \CTE{} and \CTECI{} and conclude that their utility varies depending on the domain and the value of $\delta$. 

\subsection{Simulation Study of \SDI{}, \SDICI{}, and the Bonferroni Method}

The goal of the first study was to determine how the joint prediction intervals computed by \SDI{} and \SDICI{} compare to the baseline approach of computing $d$ separate prediction intervals $[lo_j,hi_j]$ for each dimension $j=1,\ldots,d$ and applying the Bonferroni correction. Specifically, the Bonferroni baseline first computes a $1-\delta/d$ conformal prediction interval $[lo(j),hi(j)]$ by applying \SDI{} separately for each of the $d$ dimensions $j=1,\ldots,d$. These are then concatenated to create the vector prediction interval $[\lo,\hi]$. If each of the individual intervals has coverage $1-\delta/d$, then the vector prediction interval is guaranteed to achieve coverage $1-\delta$. 

\begin{figure}
    \centering
    \includegraphics[width=2.6in]{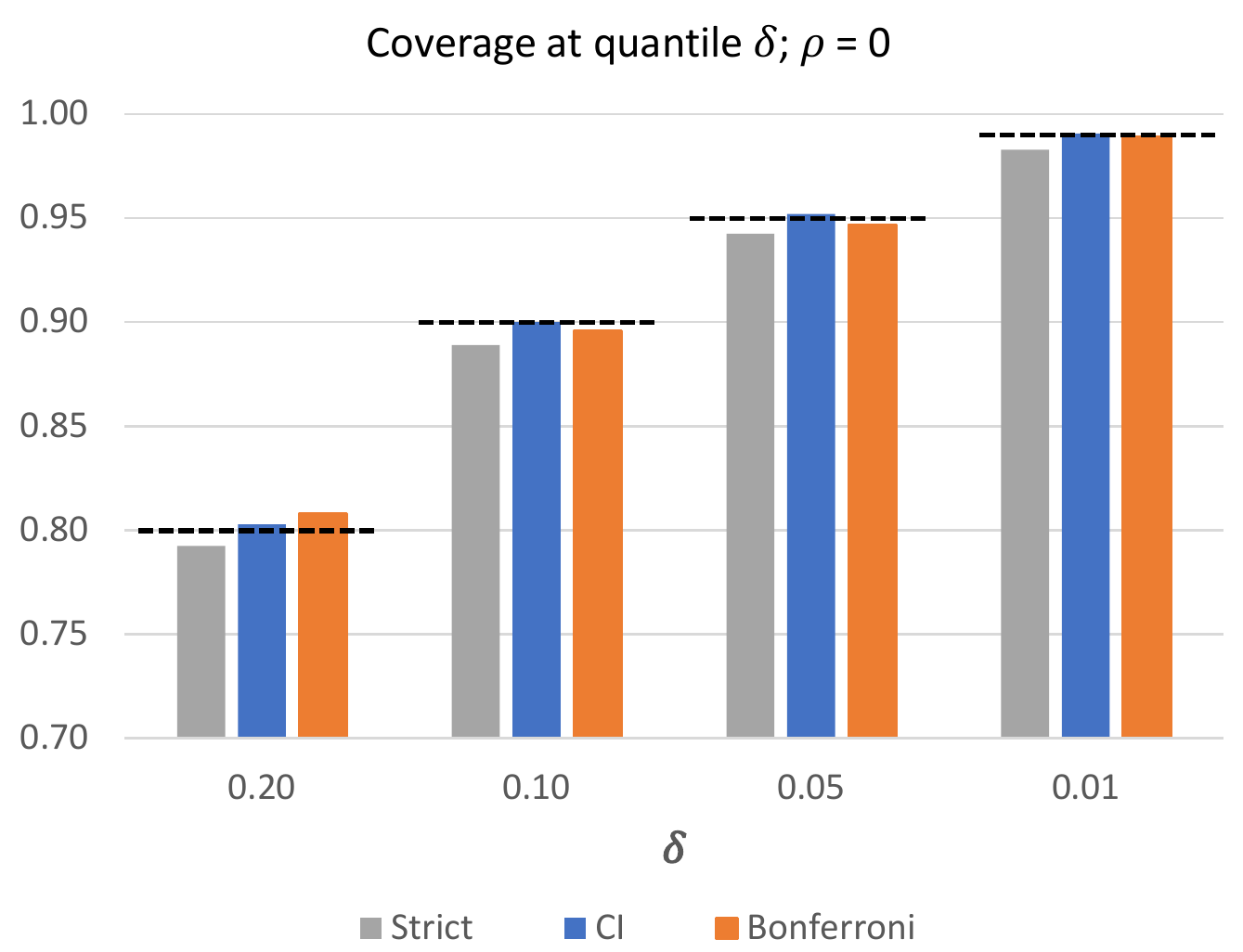} \hspace{0.2in}
    \includegraphics[width=2.6in]{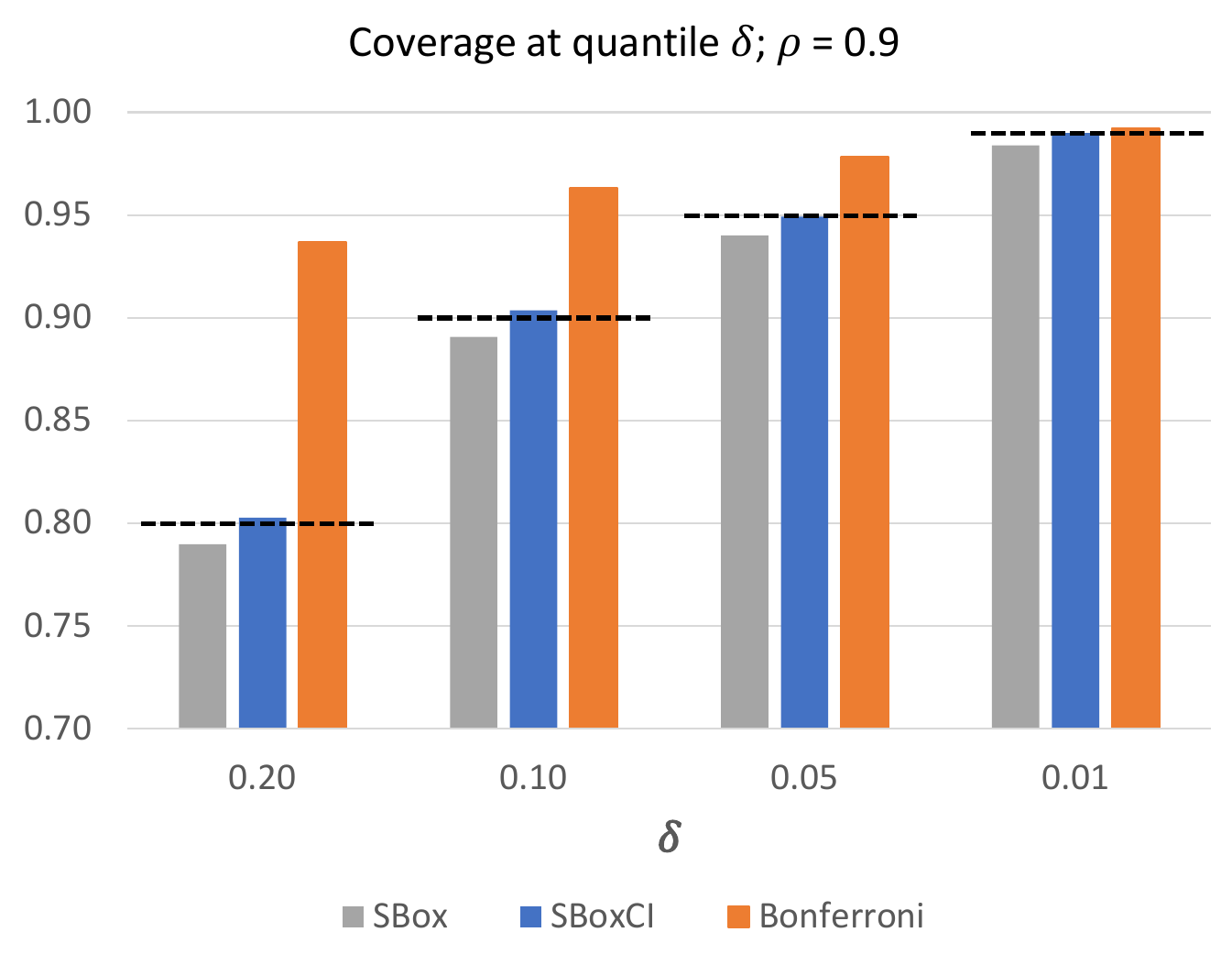}
    \caption{Comparison of prediction interval coverage for \SDI{}, \SDICI{}, and 10 separate \SDI{} intervals with Bonferroni correction for four levels of $\delta$. The dashed lines show the $1-\delta$ target coverage. Left: Correlation $\rho=0$. Right: Correlation $\rho=0.9$.}
    \label{fig:bonferroni}
\end{figure}

To compare \SDI{}, \SDICI{}, and \SDI(Bonferroni), we sampled 2000 training data points from a 10-dimensional zero-mean multivariate Gaussian with the following covariance matrix $\Sigma$. We set the diagonal elements of $\Sigma$ to 1 and all of the off-diagonal elements to $\rho$. Hence, each pair of dimensions has a correlation coefficient of $\rho$. In all three methods, we used the first 50 data points to estimate the means and standard deviations of the training data. The remaining 1950 points were employed to estimate the prediction intervals. We then evaluated coverage on a test set of 5000 data points. We conducted 100 replications of the computation. Each replication measures the fraction of test data points contained in the prediction interval. We then sorted these values and measured the $\delta$ quantile. We want our algorithms to achieve coverage of $1-\delta$ with probability $1-\delta$, which means we want the $\delta$ quantile of the coverage to be at least $1-\delta$. We also measured the mean width of the prediction intervals: $(1/d) \sum_{j=1}^d hi(j) - lo(j)$. 

Figure~\ref{fig:bonferroni} compares the results for $\rho=0.0$ and $\rho=0.9$. We observe that the $\delta$ quantile of the \SDI{} coverage is consistently below the target level of $1-\delta$. (In contrast, consistent with Proposition~\ref{prop:cp}, the mean coverage for \SDI{} is nearly perfect: 0.800, 0.899, 0.950, and 0.990 for $\rho=0$ and 0.800, 0.901, 0.950, and 0.990 for $\rho=0.9$, not shown in figure.) The upper confidence interval allows \SDICI{} to achieve the target coverage in all cases. For $\rho=0$, \SDI(Bonferroni) comes very close (computed values are 0.808, 0.896, 0.947, and 0.989), which is what we would expect, since the Bonferroni method assumes that violations of the prediction intervals will be independent in each dimension. However, for $\rho=0.9$, the Bonferroni correction is not able to take advantage of the correlations among the 10 dimensions, so the coverage at quantile $\delta$ is 0.937, 0.963, 0.978, and 0.992, which are all larger than their target values. These intervals are systematically wider. Specifically, for $\rho=0.9$ and $\delta=0.2$, the \SDI(Bonferroni) intervals are 33\% wider than the \SDI{} intervals and 30\% wider than the \SDICI{} intervals. For $\delta=0.99$, the \SDI(Bonferroni) intervals are 16\% wider than the \SDI{} intervals and 7\% wider than the \SDICI{} intervals. This simple experiment demonstrates that the \SDI{} and \SDICI{} methods are able to exploit the correlations among the dimensions much more effectively than the \SDI(Bonferroni) method.

\subsection{Studies of \SDT{} and \SDTCI{}}

In the following experiments, we compare five methods for constructing trajectory-wise prediction intervals.
\begin{itemize}
    \item Simple quantile regression (QR): We construct prediction intervals using the quantile functions fitted by \SDT{} using $\delta' = \delta$ without any conformal correction.
    \item Scaled quantile box (\SDT{}).
    \item Scaled quantile box with quantile confidence intervals (\SDTCI{}).
    \item Total exceedance (\CTE{}).
    \item Total exceedance with quantile confidence intervals (\CTECI{}).
\end{itemize}
We evaluate these methods in two Markov Decision Problems.

\subsubsection{Tamarisk}\label{sec:tamarisk}

\begin{wrapfigure}{R}{1.8in}
    \centering
    \includegraphics[width=1in]{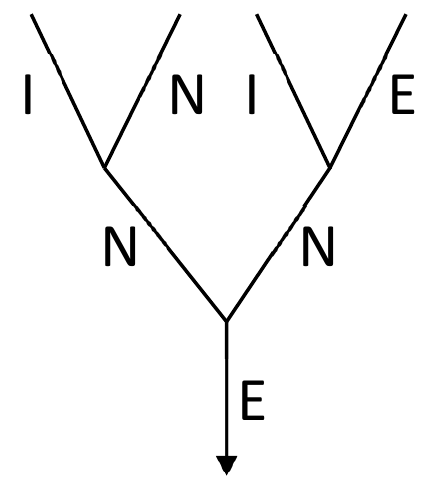}
    \caption{River network.}
    \label{fig:tamarisk-river-network}
\end{wrapfigure}

Our first evaluation domain is the problem of managing the tamarisk invasive plant species. Tamarisk has invaded the rivers of the Inter-Mountain West in the United States since the late 19th century \citep{Everitt1998}, and it has been subject to control efforts for many years \citep[e.g.,][]{Sudbrock1993}. We study a stylized version of the problem introduced by \cite{Hall2018} that consists of a river network with 7 edges arranged in a balanced binary tree (see Figure~\ref{fig:tamarisk-river-network}).

Each edge can be in one of three states: Empty, Invaded (occupied by a tamarisk tree), or Native (occupied by a native tree). The actions of the MDP consist of a 7-element vector of primitive actions, one taken in each edge of the river network. The primitive actions are Do Nothing, Eradicate (kill the tamarisk if it is present), Plant (plant a native if the edge is empty), and Eradicate+Plant (eradicate and plant). Each primitive action has a cost, and there is also a cost at each time step for each edge that is invaded. There is a budget constraint on the total action cost at each time step. The starting states are selected by sampling the state of each edge uniformly from \{Empty, Invaded, Native\}. 
\begin{figure*}
    \centering
    \includegraphics[width=\columnwidth]{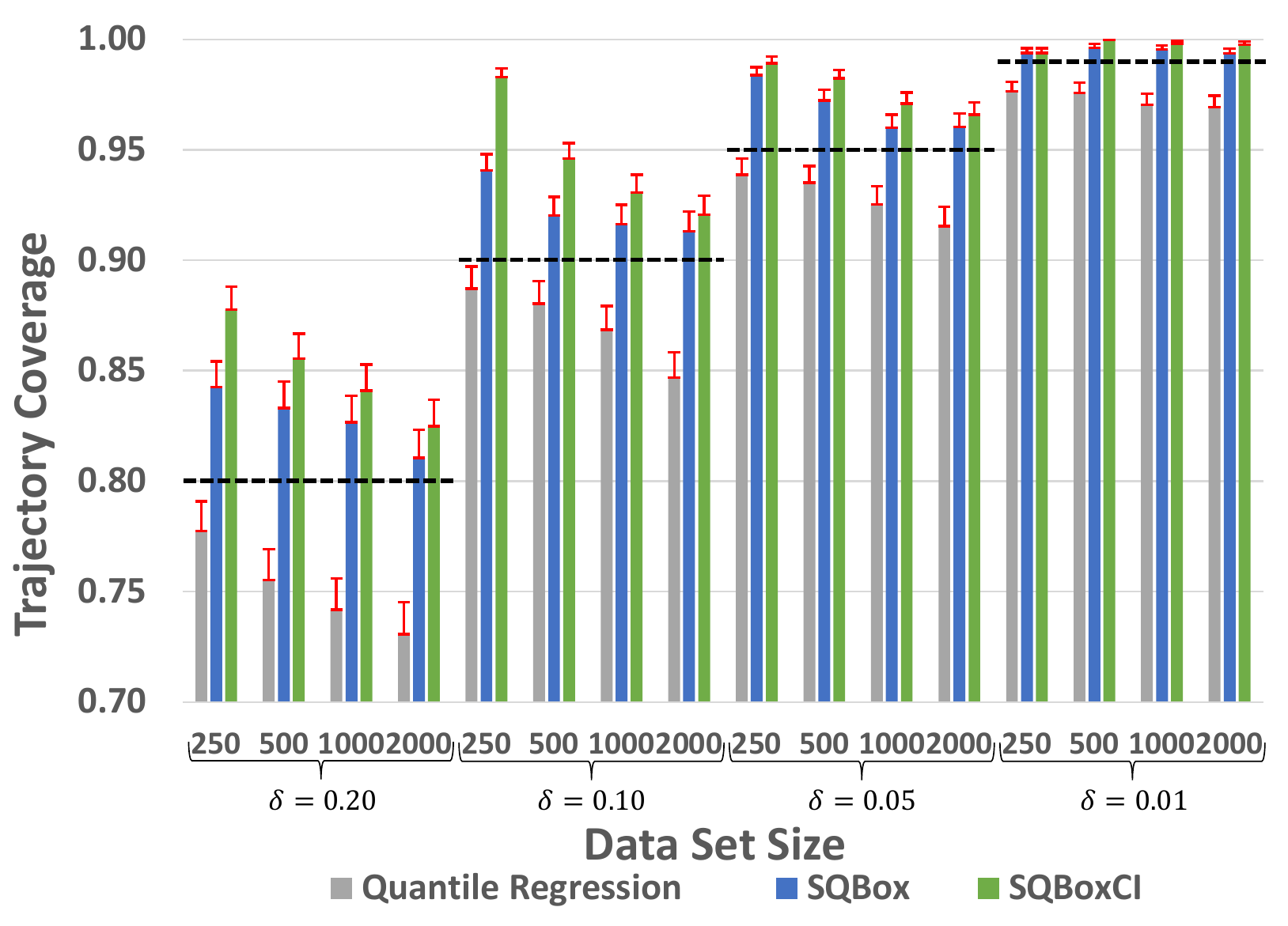}
    \caption{Fraction of Tamarisk test trajectories covered by the conformal prediction intervals with upper 99\% confidence bound. Black dashed lines indicate the target coverage $1-\delta$. Quantile Regression results (grey), intervals computed by \SDT{} (blue), and intervals computed by \SDTCI{} (green) are shown.}
    \label{fig:tamarisk-coverage}
\end{figure*}

We implemented a fixed policy based on \cite{Hall2018} that divides the river network into three levels (top, middle, and bottom) depending on the position of the edges in the tree. It computes the set of all action vectors that satisfy the budget constraint and then applies the following filters in order until only one action remains. If there are empty edges in the middle level, it plants natives (i.e., it eliminates all action vectors that do not plant natives in the middle level). If there are invaded edges at the top of the tree, it eradicates those. If the bottom edge is empty, it plants a native there. If the middle is invaded, it performs eradicate+plant. Finally, if the bottom is invaded, it performs eradicate+plant. This policy seeks to fill downstream empty slots with native plants to prevent tamarisk from becoming established in new locations. At the same time, it seeks to eliminate upstream tamarisk plants, because their seeds can spread the most effectively.

We simulated 9000 trajectories, each for 50 time steps. These trajectories were then randomly partitioned into a training set for fitting the quantile regressions, a calibration set for computing the exceedances and their prediction intervals, and a test set for evaluating the coverage of the intervals. We varied the size of the training and calibration sets and the desired value of $\delta$. The test set contained 5000 trajectories. Quantile regression was computed using Quantile Random Forests \citep{Meinshausen2006} with 1000 trees and a minimum of 20 points per leaf. To compute $\hat{\sigma}$, we employed the first 100 trajectories of the calibration set. For \SDT{} and \SDTCI{}, $\delta'=0.2$. 

\begin{figure*}
    \centering
    \includegraphics[width=2.22in]{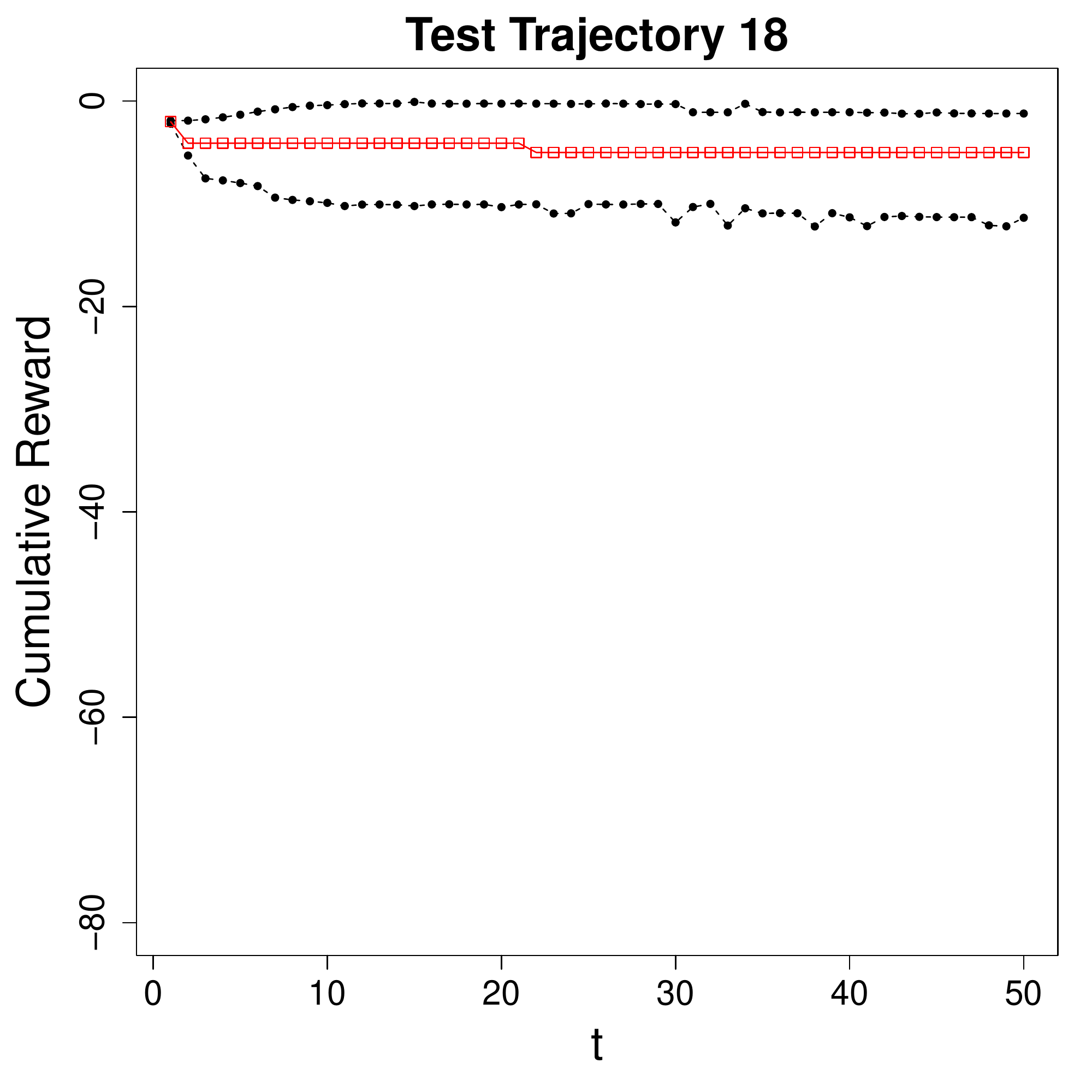}
    \includegraphics[width=2.22in]{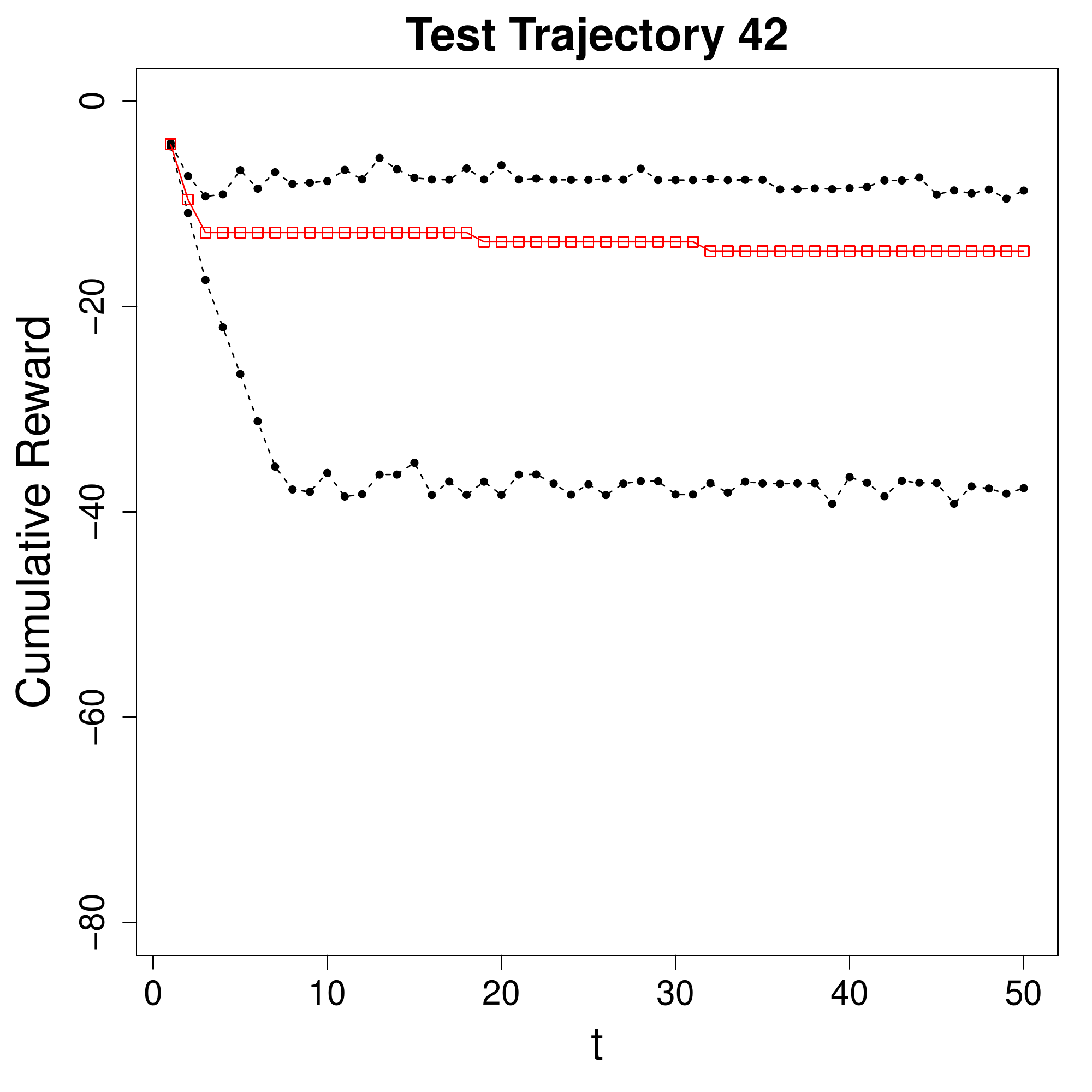}
    \includegraphics[width=2.22in]{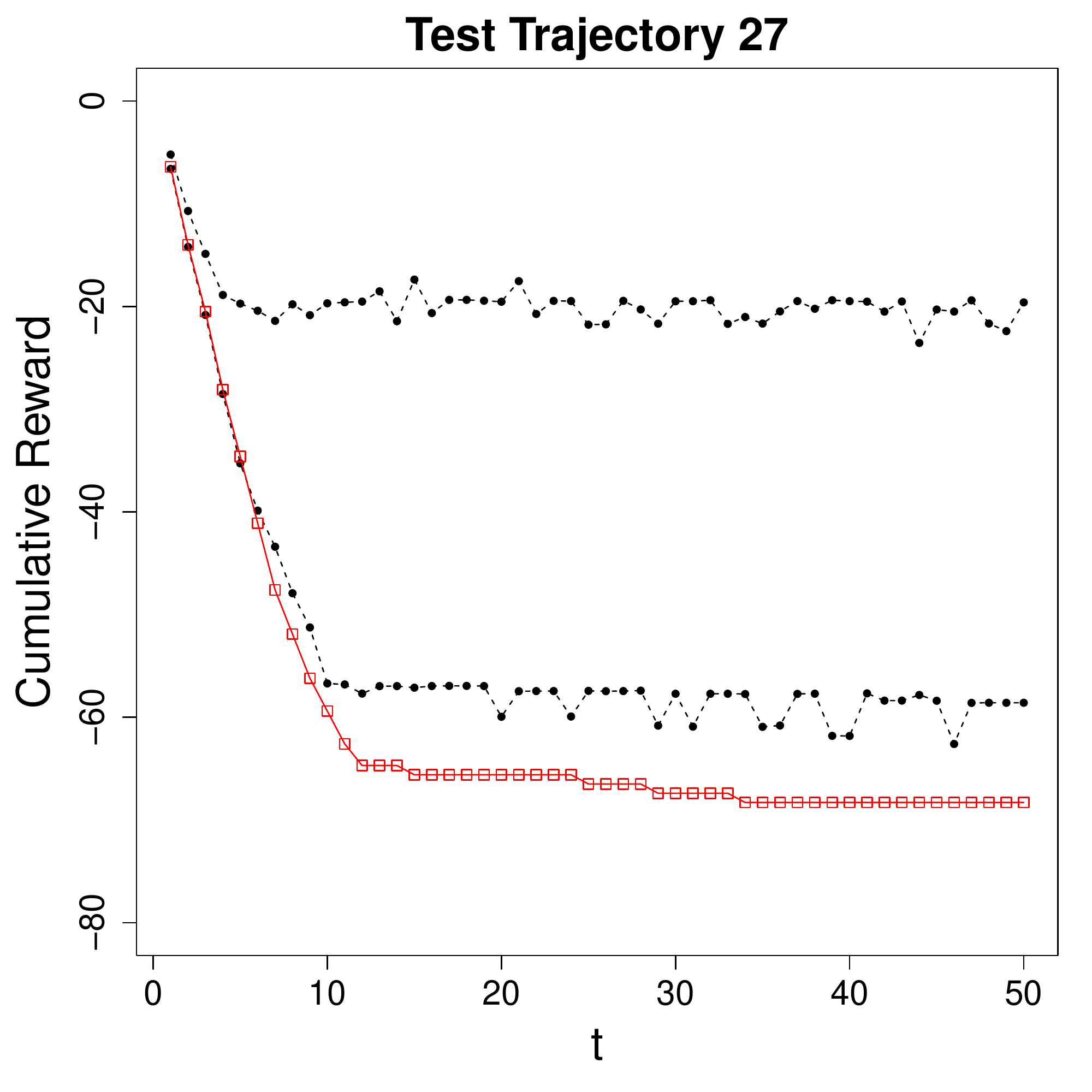}
    \caption{Prospective trajectory-wise 90\% prediction intervals (using \SDTCI{}, $n=n'=2000$, $m=100$) for three Tamarisk starting states. Black: upper and lower prediction bounds. Red: actual trajectory.}
    \label{fig:tamarisk-trajectories}
\end{figure*}
\begin{figure}
    \centering
    \includegraphics[width=\columnwidth]{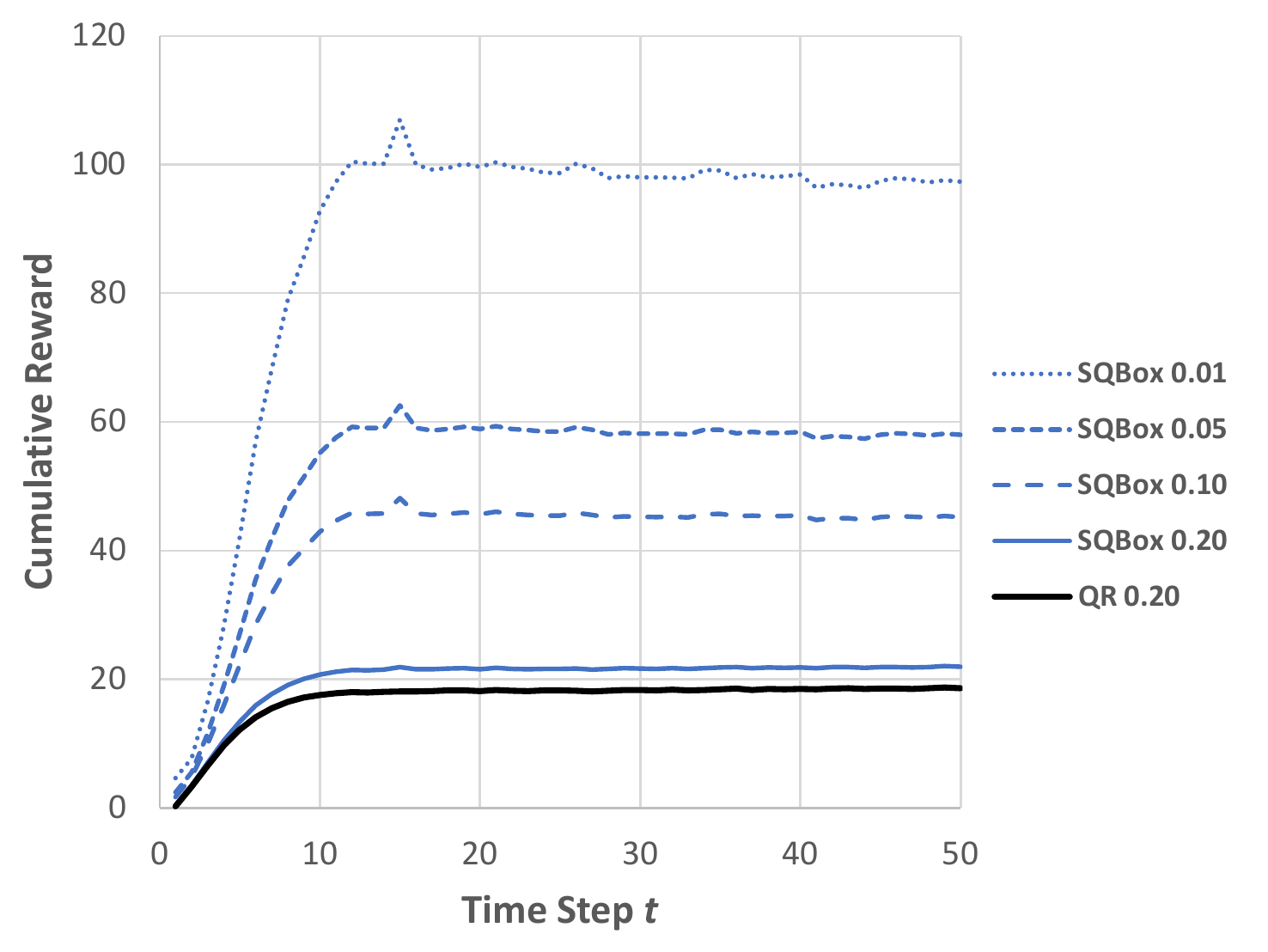}
    \caption{Mean width of prediction intervals for Tamarisk as a function of time step for $\delta \in \{0.20, 0.10, 0.05, 0.01\}$. The black line shows the widths of the quantile regressions, and the blue lines show the \SDT{} bounds. The gap between them is the conformal adjustment. Training and calibration dataset size 2000.}
    \label{fig:tamarisk-width}
\end{figure}

Figure~\ref{fig:tamarisk-coverage} summarizes the results. We say that a method achieves the target coverage of $1-\delta$ empirically if a one-sided 99\% confidence interval (based on the 5000 test trajectories) exceeds $1 - \delta$. We include this confidence bound because even with 5000 test trajectories, there is still substantial measurement uncertainty about the true coverage of the trajectory-wise conformal prediction intervals. 
By this measure, the two conformal methods (\SDT{} and \SDTCI{}) achieve the target coverage in all 16 cases. Because the \SDT{} intervals are narrower than the \SDTCI{} intervals, \SDT{} is the preferred method in this domain. Note that the raw quantile regression fails to achieve the target coverage in all cases. 

There is a trend that the intervals become tighter as the sample size increases. We believe this has two, closely-related, causes. First, when fitting the Quantile Regression Forests (QRFs), we kept the size of the leaf nodes in the QRF trees constant. As the amount of data increases, the leaves capture less variability, so the intervals shrink. Second, when computing the conformal adjustment, \SDT{} chooses the $\lceil(1-\delta)(n-l-m+1)\rceil$ element in the sorted list of $c$ values. When the number of data points is small, the gap between $(1-\delta)(n-l-m+1)$ and $\lceil(1-\delta)(n-l-m+1)\rceil$ can be quite large, whereas when $n-l-m+1$ becomes large, this gap shrinks toward zero. 

\begin{figure*}
    \centering
    \includegraphics[width=4in]{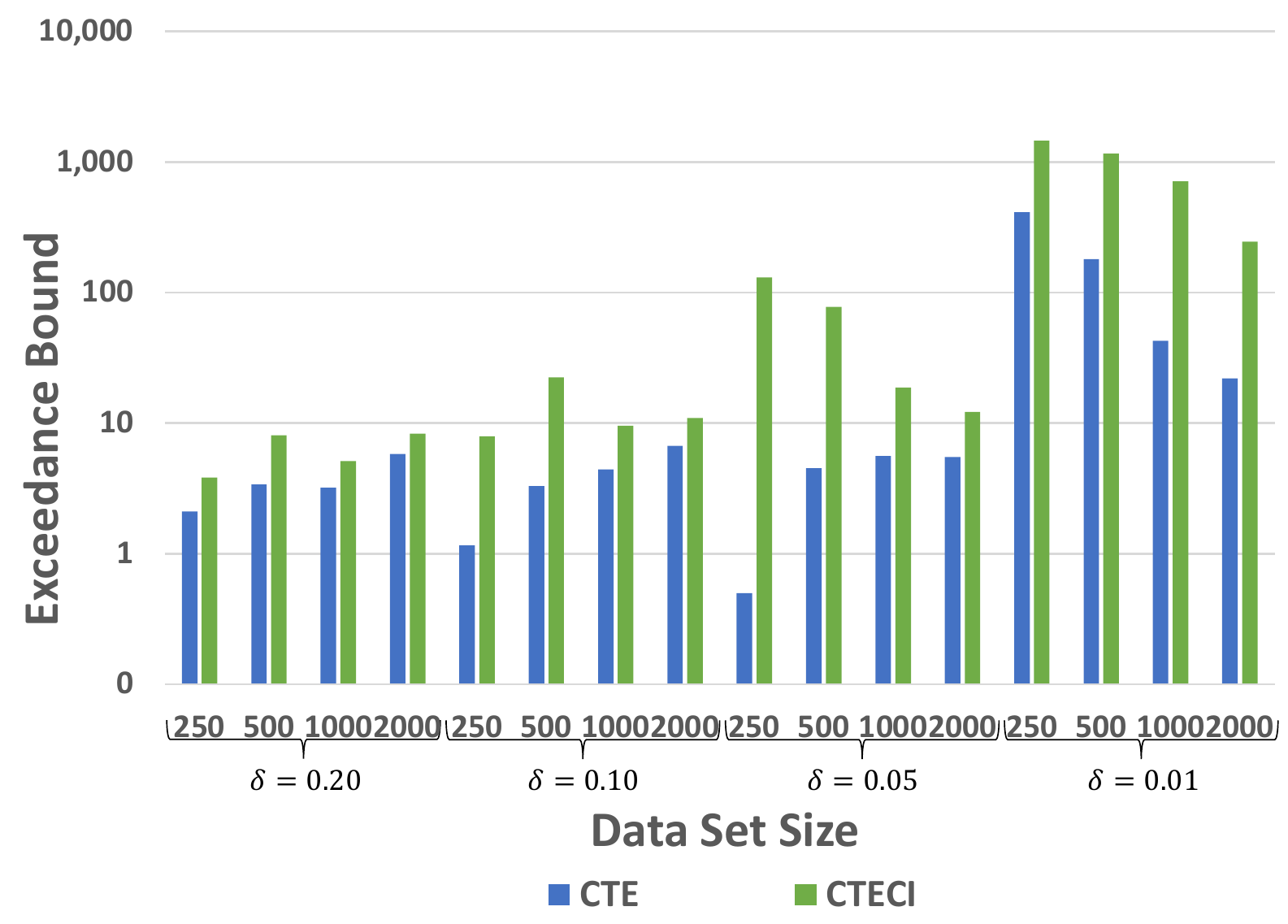}
    \includegraphics[width=4in]{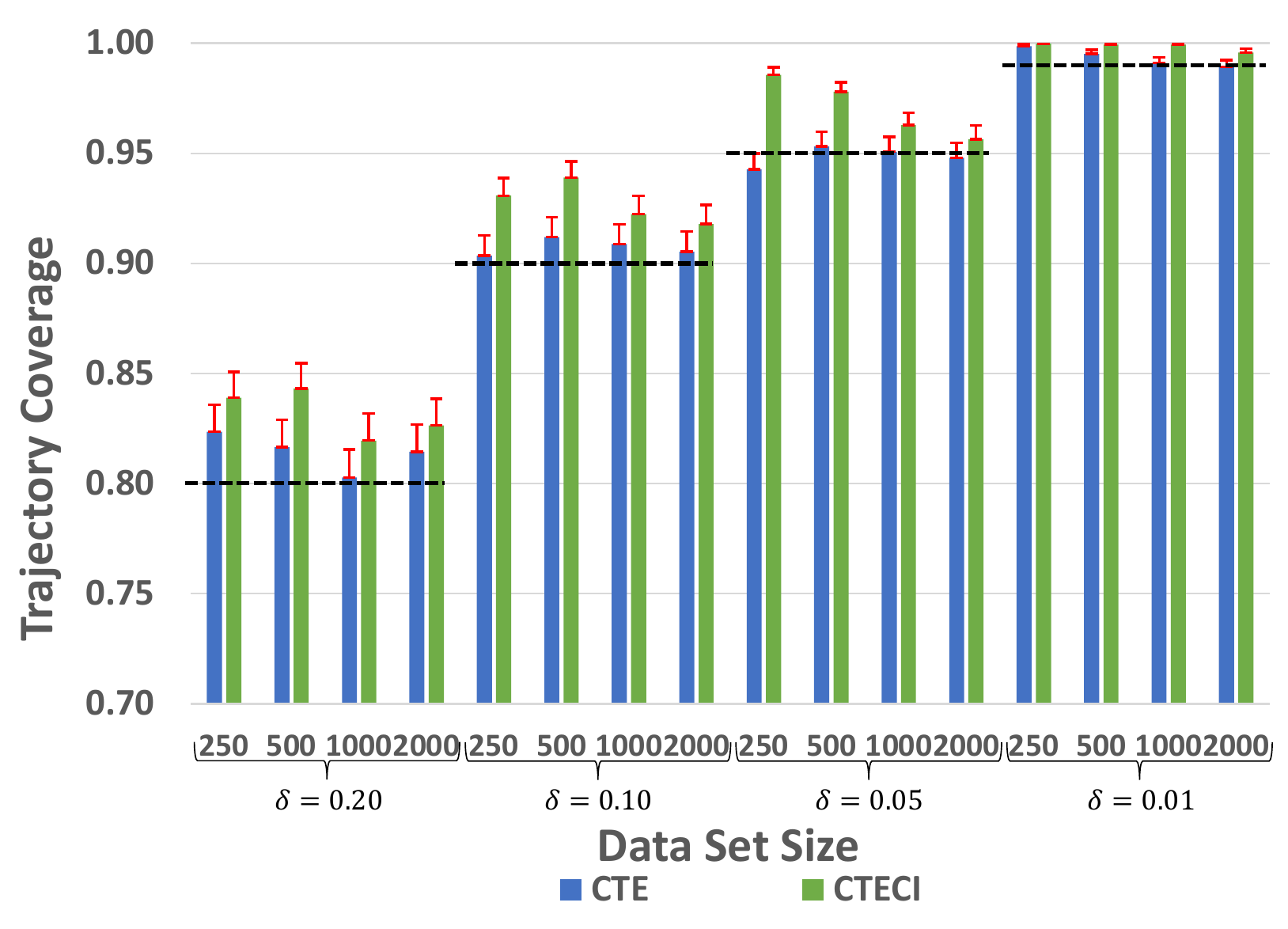}
    \caption{Upper: Conformal bounds on total exceedances for Tamarisk. Lower: Coverage of these bounds (with 99\% CI).}
    \label{fig:tamarisk-exceedance}
\end{figure*}
\begin{table}
    \centering
    \caption{Tamarisk failure analysis. Each cell corresponds to a starting state and shows the percentage of trajectories initiated in that state that exceeded the \SDTCI{} bounds for $n=2000$ and $\delta = 0.1$. Grey scale encodes the magnitude. White cells are significantly larger than $\delta$ based on a one-sided exact binomial hypothesis test ($p< 0.05$).}
    \label{tab:tamarisk-failure-analysis}
    \includegraphics[width=4in]{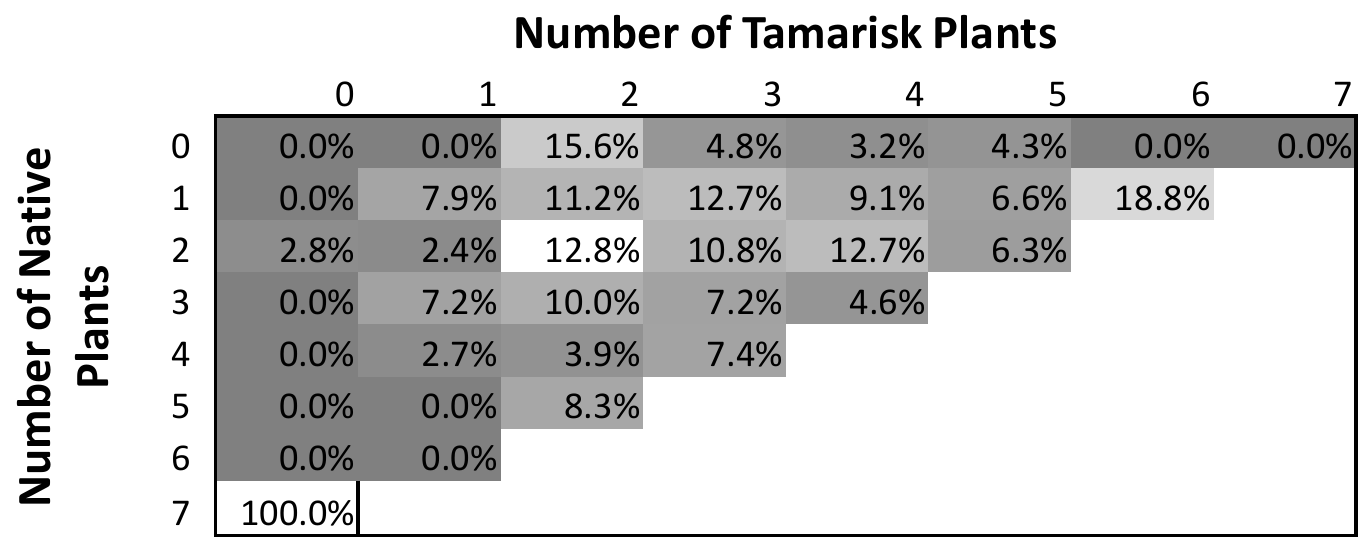}
\end{table}

Figure~\ref{fig:tamarisk-trajectories} shows prospective prediction intervals for three different starting states (for $\delta = 0.2$ and 2000 training and calibration trajectories). Trajectory 18 had 1 tamarisk and 6 native edges in the starting state. The model correctly anticipates that the total cost will be small. Trajectory 42 had 4 tamarisk and 3 native edges. The model correctly anticipates that it will be expensive to kill the tamarisk plants and replace them with natives. The wide bounds reflect the stochasticity of the reproduction and spread of the seeds. The actual trajectory was ``lucky'': at $t=1$, one tamarisk tree died and was replaced by a native via seed propagation. This happened again at $t=2$, so that instead of requiring four eradication actions, the invader was eliminated after only two. Trajectory 27 starts with 6 tamarisk and 1 native, so the model predicts an even worse range of outcomes. Despite this, the trajectory is ``unlucky'': over the course of treatment, 6 natives died and were replaced by tamarisk, and one edge had to be treated four times before the tamarisks were finally eliminated at $t=12$. This results in substantial violations of the prediction interval. Across the three cases, note how the prospective prediction bounds change depending on the starting state. This is due entirely to the quantile regressions, which are conditional. In contrast, the conformal adjustments are the same for all trajectories.

Figure~\ref{fig:tamarisk-width} plots the mean width of the prediction intervals for \SDT{}. We have also plotted the width of the $\delta'=0.2$ QR interval upon which the \SDT{} intervals are constructed. The gap between them is the conformal adjustment, and this grows rapidly as $\delta$ shrinks from 0.2 to 0.01.  

We have emphasized that the conformal prediction intervals are only semi-conditional. Hence, it is important to check whether there are regions within the state space that systematically violate the prediction intervals. Table~\ref{tab:tamarisk-failure-analysis} shows an example of this analysis. Each cell in the table corresponds to a starting state. The number of empty edges in the river graph is not shown, but it can be computed by subtracting the sum of the native and tamarisk plants from 7. Each cell shows the percentage of trials starting in that cell that violated the \SDTCI{} bounds for $n=2000$ and $\delta=0.10$. Ideally, these percentages would all be 10\%. We performed an exact one-sided binomial confidence interval at the $0.95$ confidence level to determine which of these values are likely to be greater than 10\%, and only two starting states pass this test: (7 native, 0 tamarisk) and (2 native, 2 tamarisk). Of course we would expect 1-2 false discoveries (32 independent tests at $p=0.05$).  In the case of the (7 native, 0 tamarisk), there were only three trajectories that started in this state, but all three of them violated the bounds. So it is possible that the bounds are too tight in this extremely beneficial case. However, 3 is a very small sample. In summary, there is very little evidence for local clusters of failures in this problem. This suggests that the semi-conditional prediction intervals are giving a good approximation of fully-conditional intervals on this problem.

As an alternative to bounding the trajectory at each step, Figure~\ref{fig:tamarisk-exceedance} (upper) shows the conformal bounds on total exceedances as a function of $\delta$ and data size. Note the log scale. Figure~\ref{fig:tamarisk-exceedance} (lower) shows that conformal total exceedance (with and without a confidence interval addition) attains the desired coverage in all 16 configurations. For $\delta \in \{0.20, 0.10, 0.05\}$, the conformal bounds are less than 10, which suggests that the combination of the quantile regressions and the total exceedance bound give a useful prospective picture of how the policy will behave. However, for $\delta = 0.01$, the bounds range from 22 to 414, which suggests that the trajectory-wise quantile regressions (from which the total exceedances are computed) are no longer faithfully representing policy behavior.

\subsubsection{Starcraft}
Our second evaluation domain involves simple battles in the video game \textit{Starcraft~2} \citep{vinyals2017starcraft}. A Blue team of units faces a Red team. The initial number of units for the Blue team is chosen uniformly in $\{5, \ldots, 20\}$; the initial number of units for Red is chosen uniformly in $\{5, \ldots, 10\}$. The Blue units are controlled by a fixed policy that commands all units to advance toward the Red team at $t=0$. When opposing units come within range of one another (around $t=5$), they engage in combat controlled by the internal Starcraft logic. At $t=14$, the Red team receives an additional set of units whose number is drawn uniformly from $\{0, \ldots, N\}$, with $N$ itself drawn uniformly from $\{0, \ldots, 15\}$. These reinforcements introduce substantial uncertainty into the game, and there is also some stochasticity in the behavior of individual units. Blue receives a reward of $+1$ for each Red unit that is destroyed and $-1$ for each Blue unit that is lost. As discussed below, we added zero-mean Gaussian noise with a standard deviation of 0.05 to these rewards to prevent large numbers of ties in the behavior values.

\begin{figure*}
    \centering
    \includegraphics[width=\columnwidth]{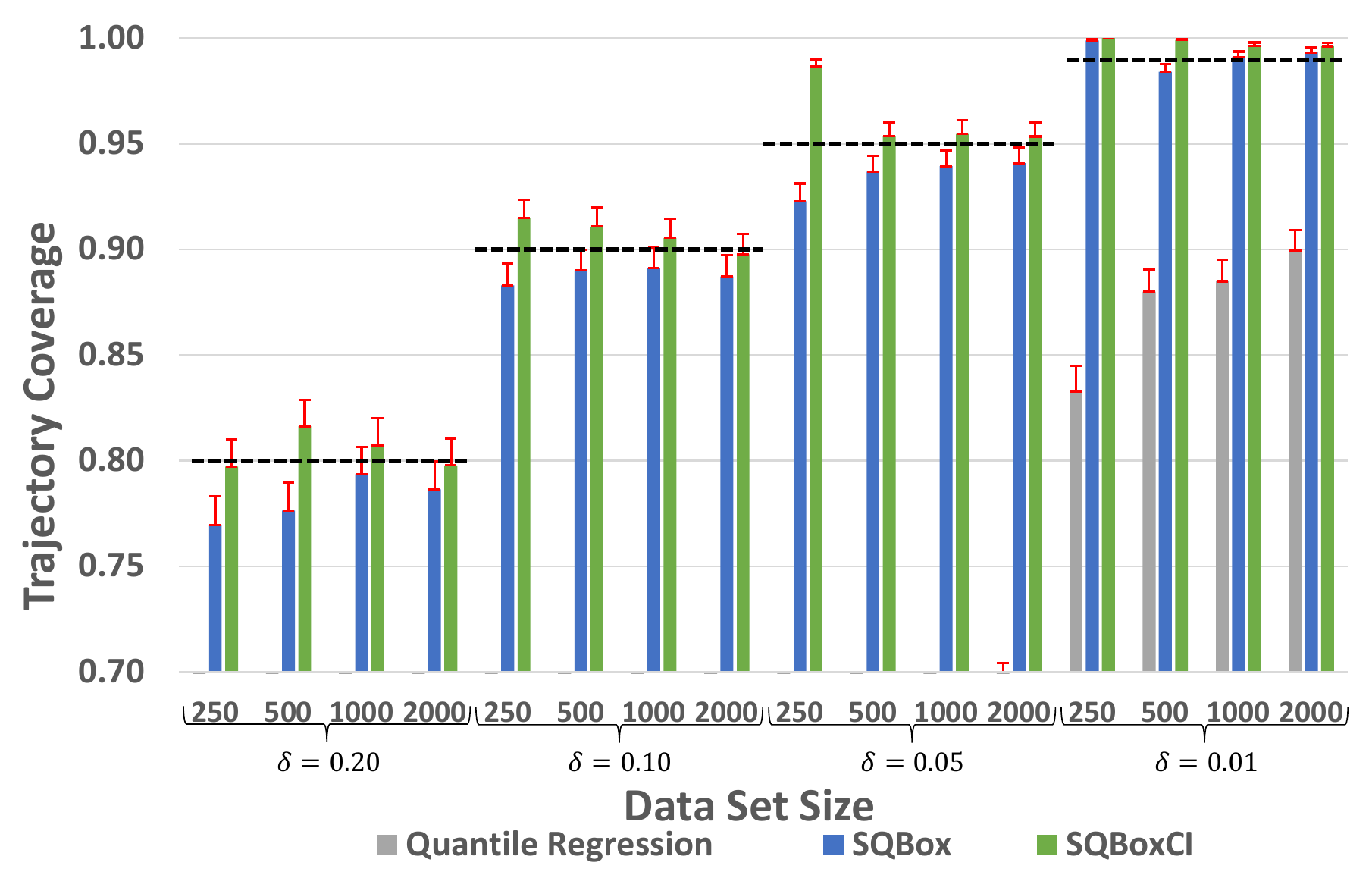}
    \caption{Fraction of Starcraft test trajectories covered by the conformal prediction intervals with 99\% upper confidence bounds. Black dashed lines indicate the target coverage $1-\delta$. Quantile regression results (``Quantile Regression'', grey), \SDT{} (blue), and \SDTCI{} (green).}
    \label{fig:starcraft-coverage}
\end{figure*}
\begin{figure*}
    \centering
    \includegraphics[width=\columnwidth]{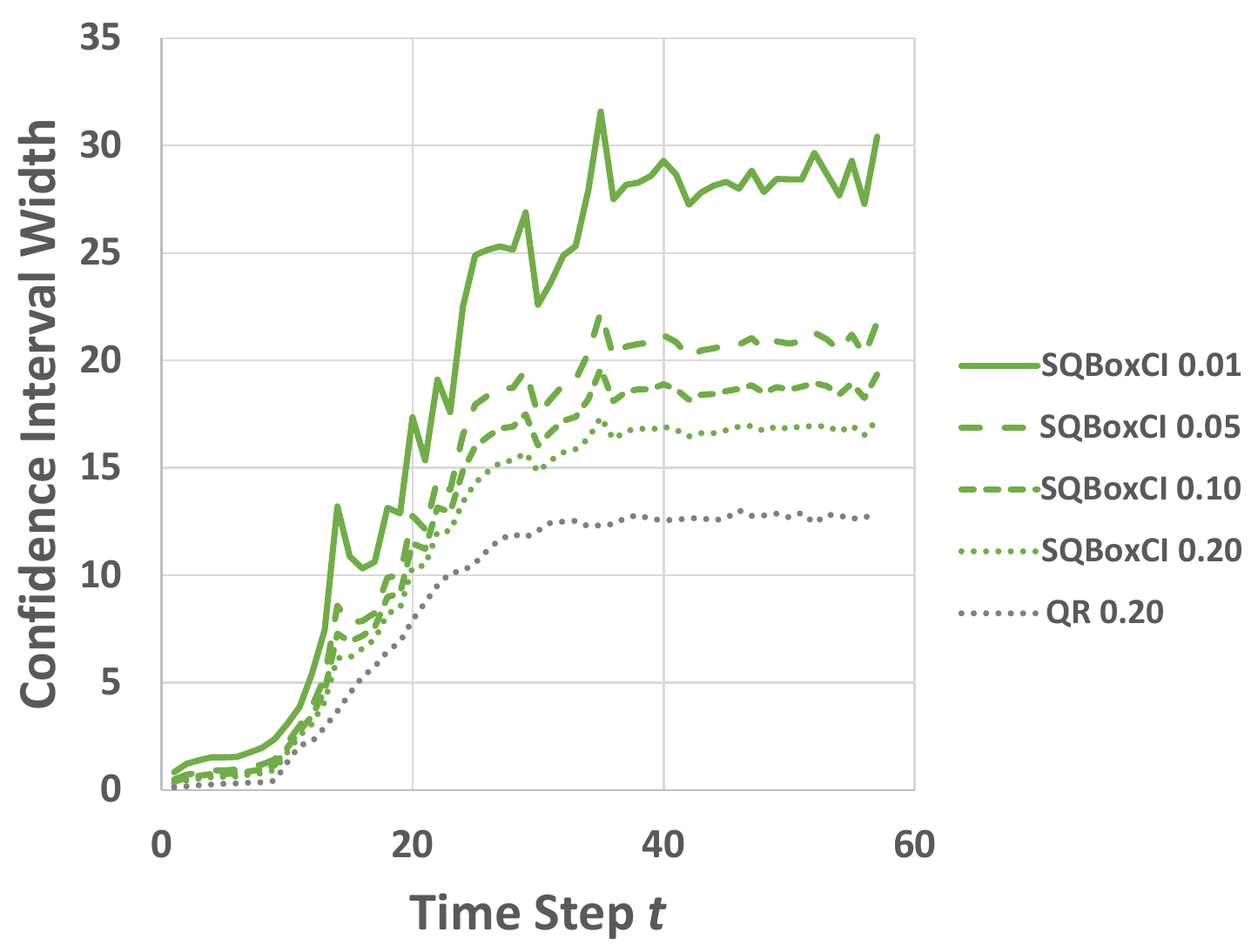}
    \caption{Mean width of \SDTCI{} prediction intervals (green) for Starcraft as a function of time step for $\delta \in \{0.20, 0.10, 0.05, 0.01\}$ along with the ``inner'' Quantile regression (QR) prediction interval (grey). Training and calibration dataset size 2000.}
    \label{fig:starcraft-width}
\end{figure*}
\begin{figure*}
    \centering
    \includegraphics[width=4in]{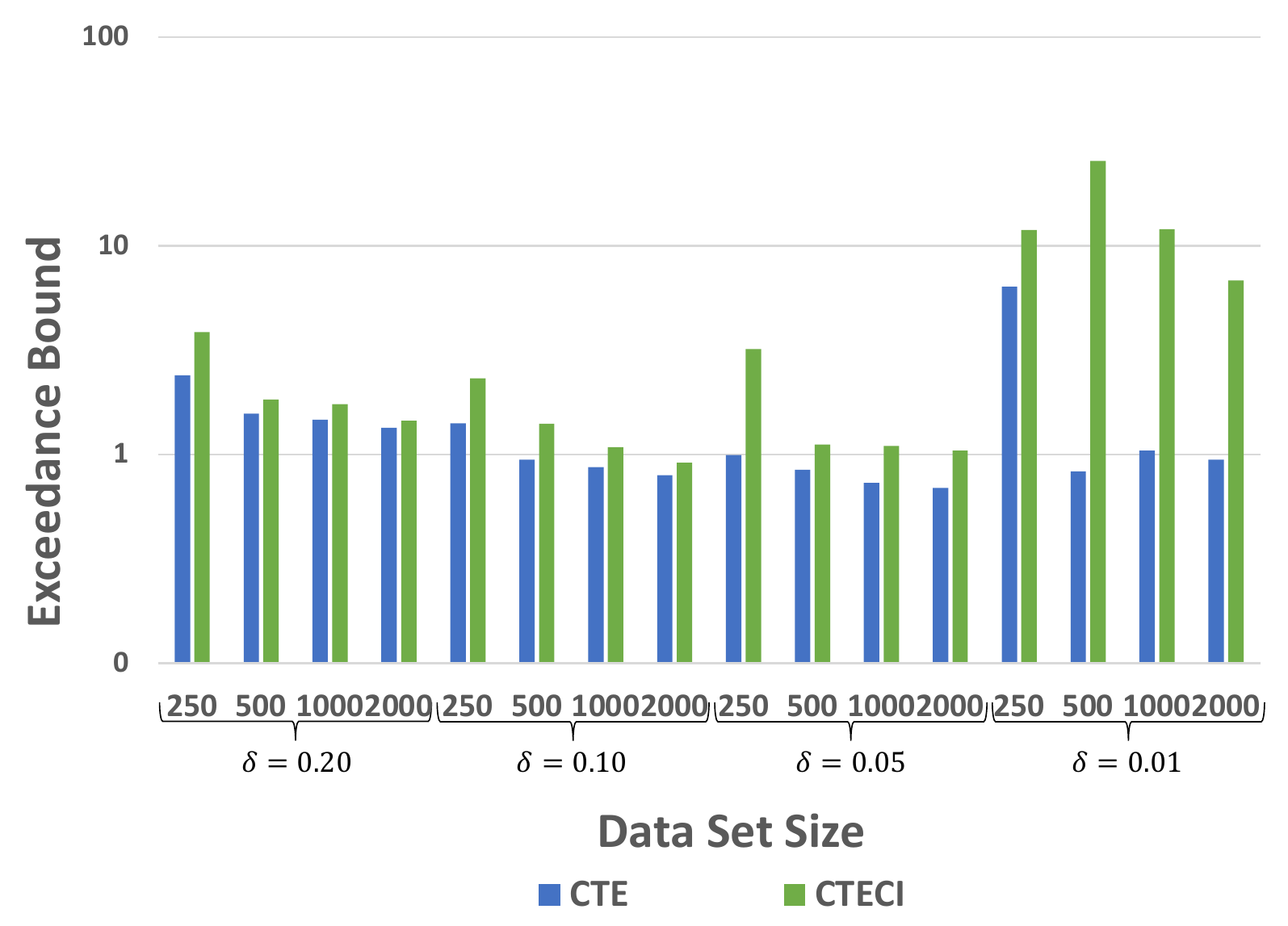}
    \includegraphics[width=4in]{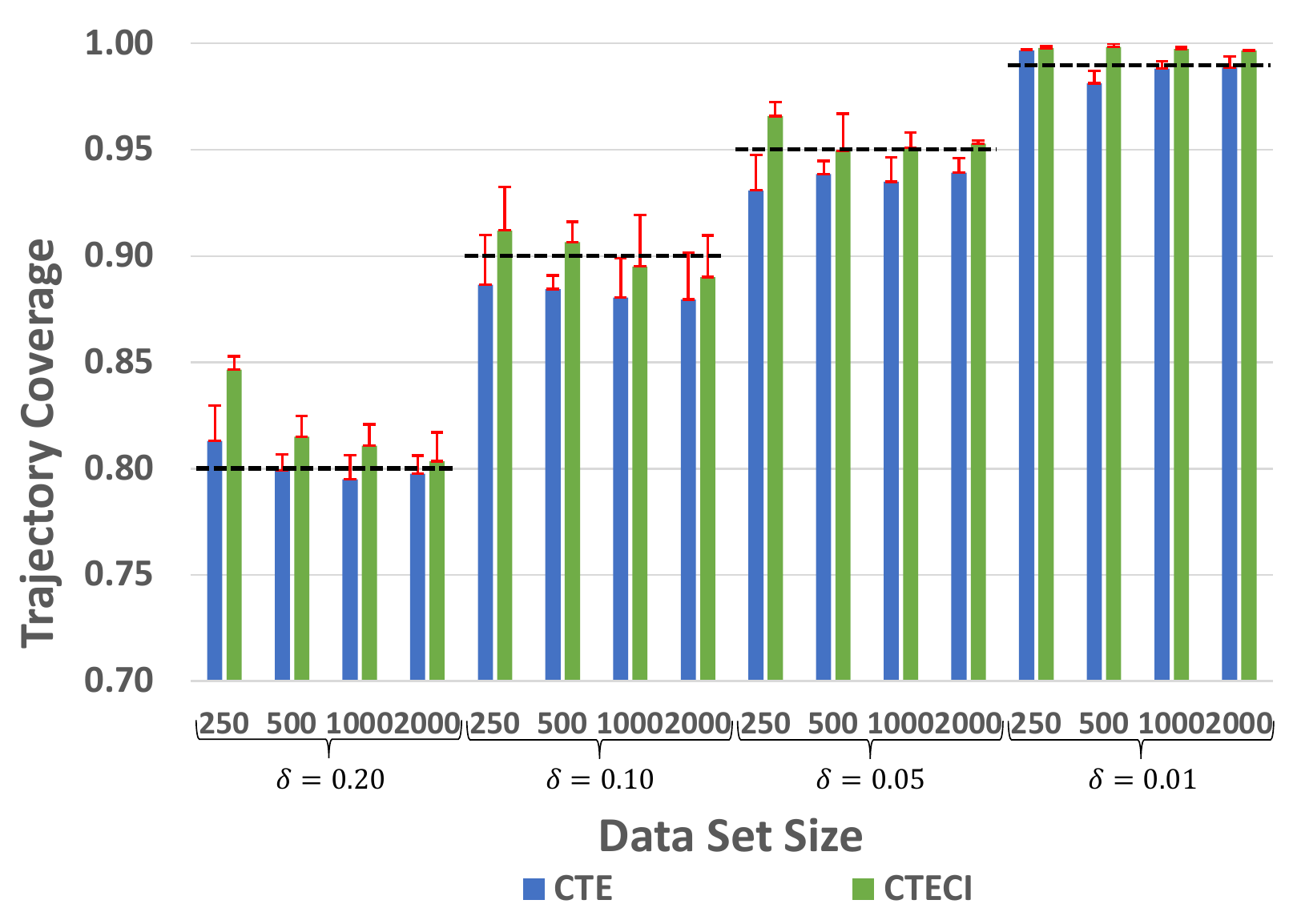}
    \caption{Upper: Conformal bounds on total exceedances for Starcraft. Lower: Coverage of these bounds (with 99\% CI).}
    \label{fig:starcraft-exceedance}
\end{figure*}
\begin{figure*}
    \centering
    \includegraphics[width=2.22in]{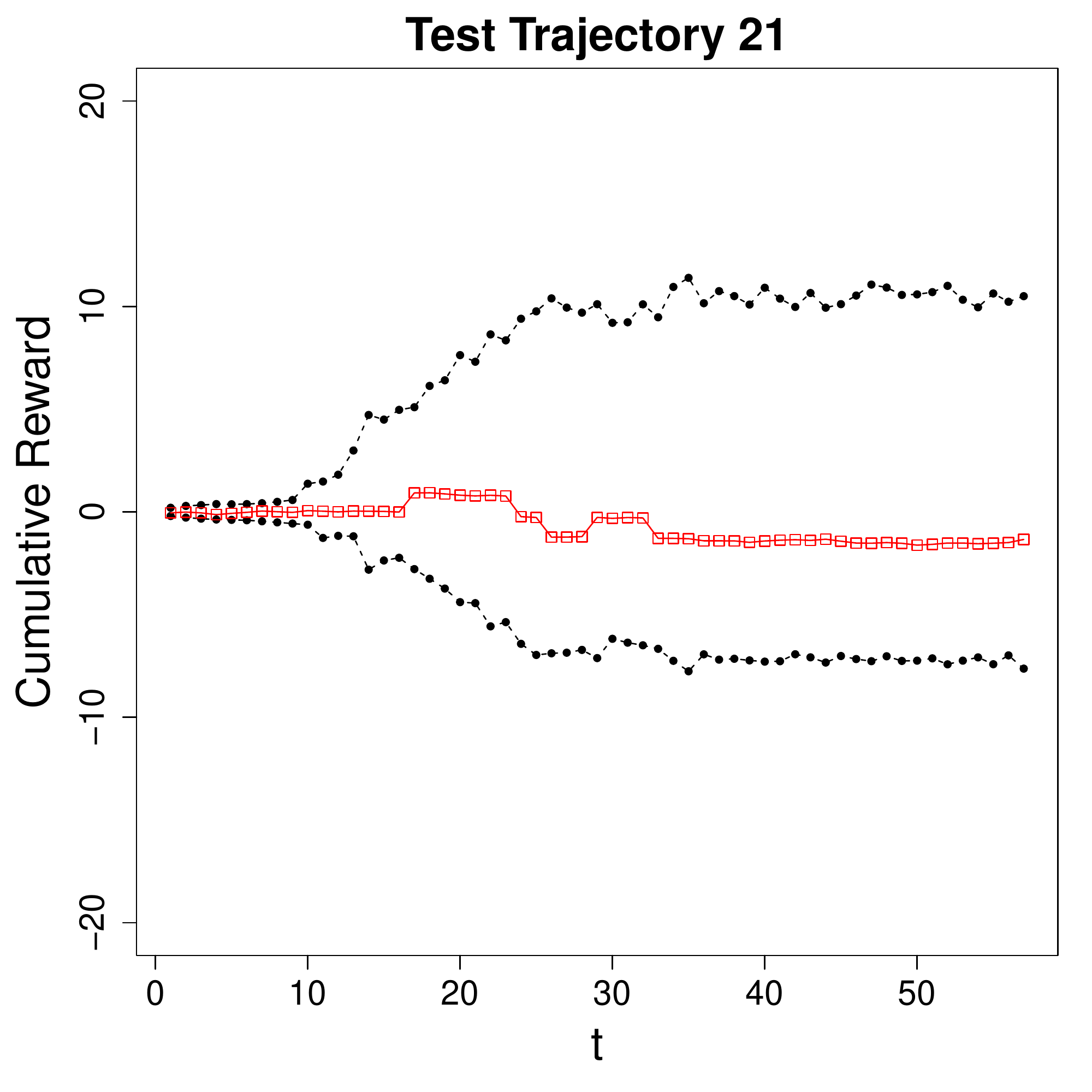}
    \includegraphics[width=2.22in]{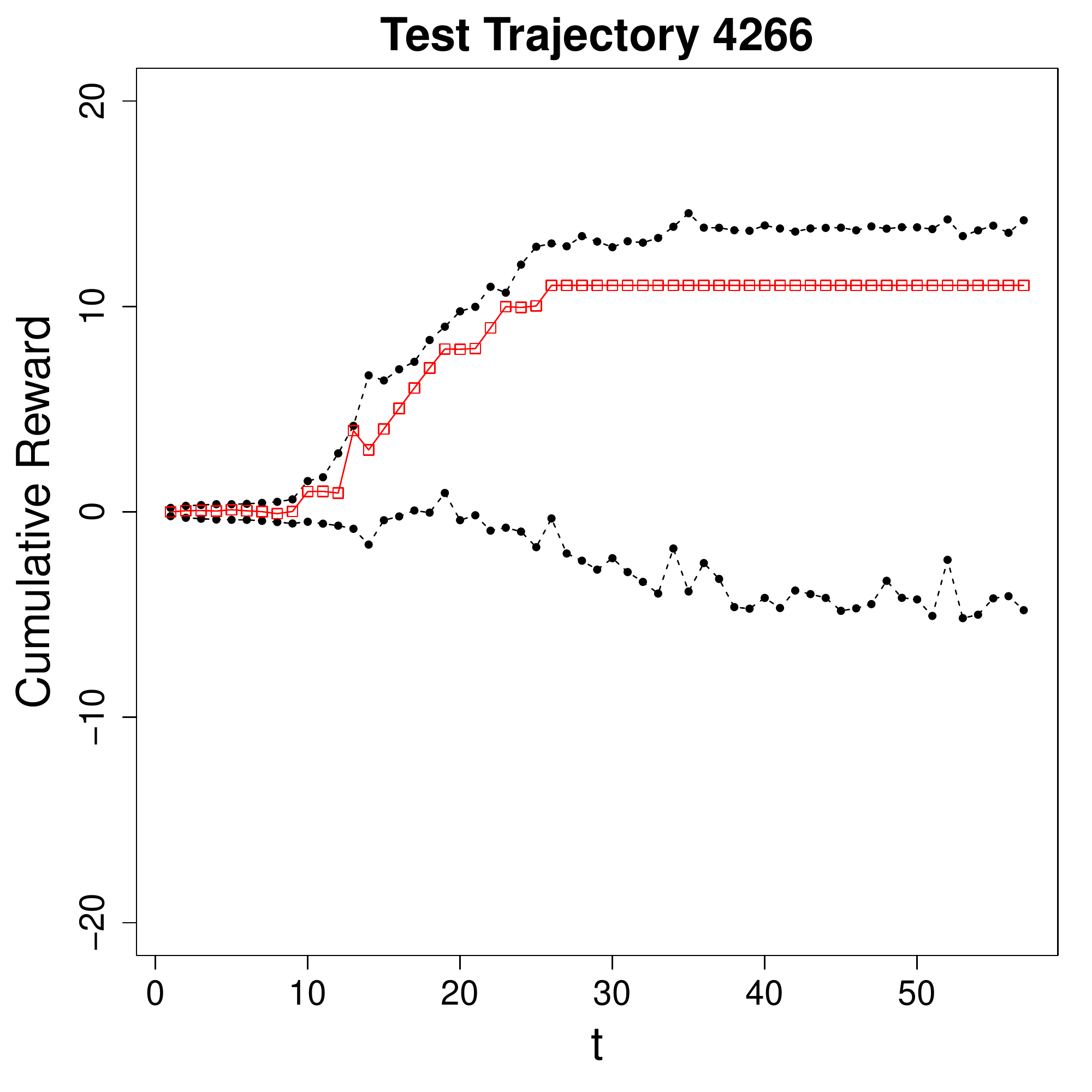}
    \includegraphics[width=2.22in]{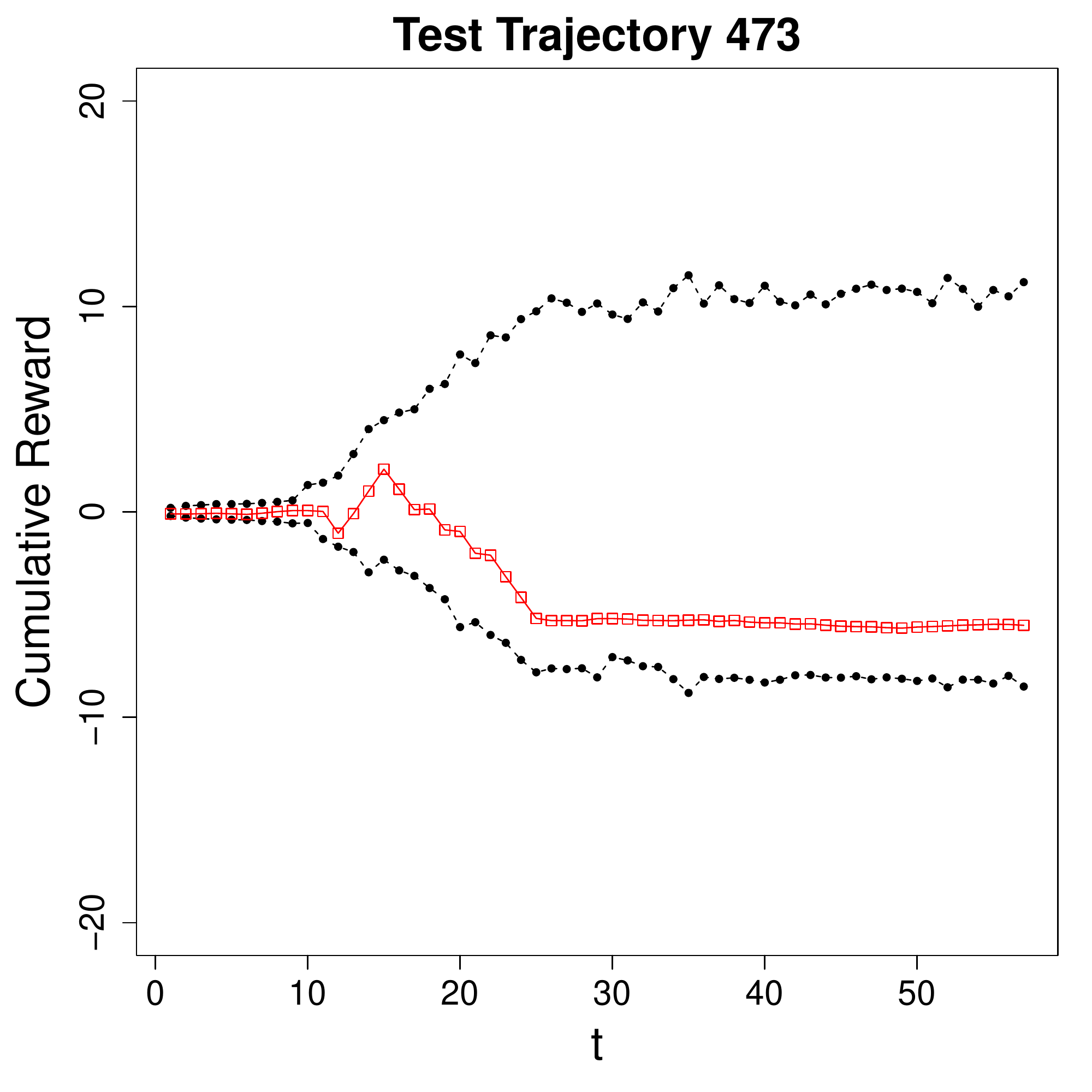}
    \caption{Prospective trajectory-wise 90\% prediction intervals (\SDTCI{}, $n=n'=2000$, $m=100$) for three Starcraft starting states. Black: upper and lower prediction bounds. Red: actual trajectory.}
    \label{fig:starcraft-trajectories}
\end{figure*}
\begin{table}
    \centering
    \caption{Starcraft failure analysis. Each cell corresponds to a starting state and shows the percentage of trajectories initiated in that state that exceeded the \SDTCI{} bounds for $n=2000$ and $\delta = 0.1$. Grey scale encodes the magnitude. White cells are significantly larger than $\delta$ based on a one-sided exact binomial hypothesis test ($p< 0.05$).}
    \label{tab:starcraft-failure-analysis}
    \includegraphics[width=2.5in]{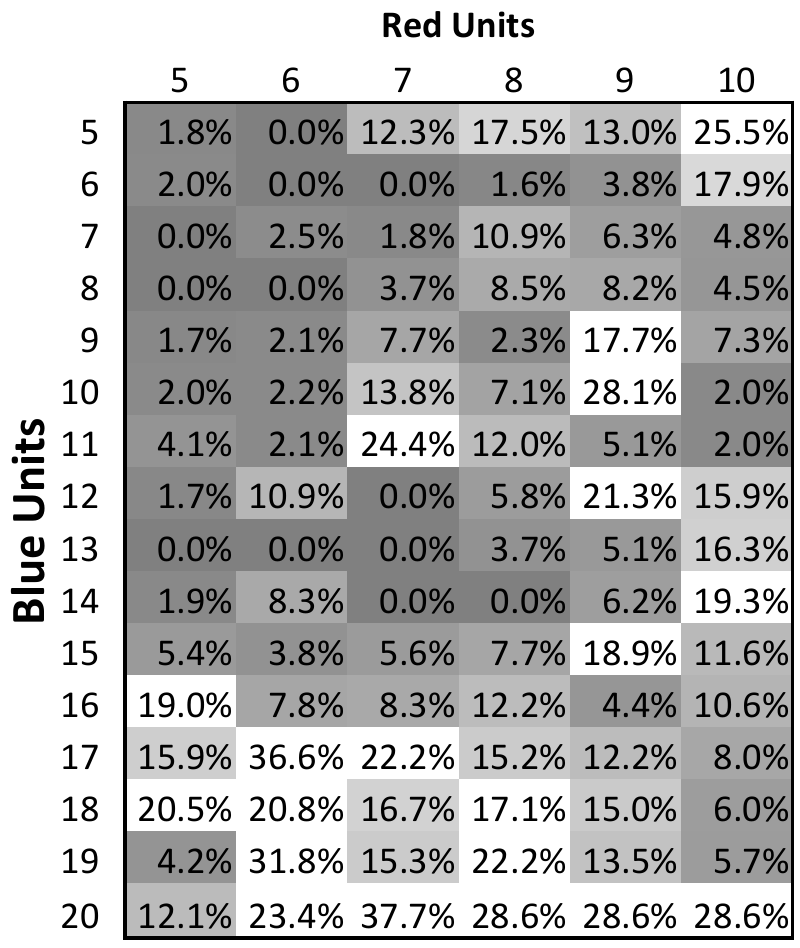}
\end{table}
As with the Tamarisk problem, we employed 5000 trajectories for testing and varied the number of training and calibration trajectories from 250 to 2000. Quantile regressions were computed with $\delta'=0.2$ using a quantile regression forest with 1000 trees and leaf size 20. Figure~\ref{fig:starcraft-coverage} shows the coverage of the prospective trajectory-wise prediction intervals. The quantile regression intervals (with $\delta'=\delta$) give extremely poor coverage and never achieve the target coverage of $1-\delta$. \SDT{} achieves the target coverage in only 6 of the 16 configurations, whereas \SDTCI{} succeeds on all 16 configurations. We again observe a trend where the prediction intervals become tighter (closer to the target coverage) as the amount of training data increases, particularly for \SDTCI{}. However, we do not observe this behavior in the quantile regressions themselves, so we suspect in this case the main factor is the fixed leaf size of the quantile regression forests. 

We discovered an interesting behavior when we performed this experiment using the discrete $\pm1$ Starcraft rewards without added Gaussian noise. In several cases, the prediction intervals for \SDT{} and \SDTCI{} were surprisingly wide, with coverages of 0.98 even for $\delta = 0.10$ and $0.05$. Theorem 2 in \cite{Lei2018} guarantees that conformal prediction intervals are tight when the underlying distribution is continuous. However, when the distribution is discrete, there can be large sets of identical $c_i$ values so that the $\lceil(1-\delta)(n-m+1)\rceil$ value is the same across many different values of $\delta$. In this case, we were sometimes obtaining the same value for $\delta=0.10$, $0.05$, and $0.01$. In this setting, \SDTCI{} still achieved the target coverage in all 16 configurations, but the prediction intervals were very wide. By adding very small amounts of noise to the Starcraft rewards, this problem of ties is eliminated, and Figure~\ref{fig:starcraft-coverage} shows that the prediction intervals are satisfyingly tight except at the smallest sample size $n=250$.

Figure~\ref{fig:starcraft-width} plots the width of the \SDTCI{} trajectory-wise prediction intervals along with the width of the quantile regression interval for $\delta'=0.2$. The differences between the \SDTCI{} curves and the quantile regression curves are the conformal adjustments. The increase in interval width due to conformal adjustments is not nearly so dramatic as in the Tamarisk domain, which reflects the lower degree of stochasticity in Starcraft.

We performed a failure analysis of the Starcraft trajectories for $n=2000$ and $\delta=0.10$, as shown in Table~\ref{tab:starcraft-failure-analysis}. Unlike in the case of Tamarisk, Starcraft exhibits a large number of starting states in which the fraction of failures is significantly larger than the target value of $0.10$. In particular, when the number of Blue units minus the number of Red units is 10 or more, there are 13 out of 21 starting states with elevated failure rates. Hence, in this region, at least, the \SDTCI{} prediction intervals are not valid. More generally, it is not safe to interpret the \SDTCI{} prediction intervals in Starcraft as fully conditional. 

When a region of failed states is detected, one remedy is to decompose the original problem into two regions and compute conformal adjustments separately for starting states in each region.  There is no need to stop with two regions: If sufficient calibration data are available, we can progressively decompose a problem into multiple regions and produce semi-conditional prediction intervals within each region that give a good approximation to fully-conditional prediction intervals. \cite{Lei2013a} pursue a related idea by fitting a functional Gaussian mixture model and then produce separate prediction bands for each mixture component.

As in the Tamarisk problem, we explored the alternative of providing a guarantee based on the total exceedance. Figure~\ref{fig:starcraft-exceedance} plots the total exceedance bounds and coverage. The \CTE{} method achieves the target coverage in 9 cases, whereas \CTECI{} succeeds in all 16 cases. Focusing on the latter, the total exceedance bounds are very tight. Most are in the 1-3 range except for $\delta = 0.01$ where the maximum is 25. Hence, we can see that quantile regression combined with a conformalized bound on total exceedance gives a good characterization of the behavior of the Starcraft policy.

Figure~\ref{fig:starcraft-trajectories} shows a range of different Starcraft starting states, prediction intervals, and actual behavior. In the starting state for trajectory 21, the Blue team has 6 units and the Red team has 5. This near balance leads to a prediction interval that ranges from $-6$ to $+12$. In this trial, Red received only two reinforcement units, so the cumulative reward is almost flat. In the starting state of trajectory 4266, Blue has 18 units and Red has 6, and consequently the prediction interval ranges from $-6$ to $+14$. Notice that the lower bound increases slightly at $t=19$, so it is expecting that Blue will not be hurt by the Red reinforcements. In this case, Blue attains a decisive win, even though Red receives 6 reinforcement units at $t=14$. Finally, trajectory 473 starts with both Blue and Red having 9 units, so it is also well balanced. However, the prediction interval is slightly wider than for trajectory 21: $-7$ to $+13$. It seems that the larger unit count leads to slightly more uncertainty in the outcome of the battle. In this particular trajectory, Red receives 4 units of reinforcements, and we observe an upward bump in the trajectory at $t=14$ as Blue is able to destroy some of these new units. However, the battle ends up going poorly for Blue, and it loses to Red at around $t=26$. 

\section{Concluding Remarks}

This paper introduced a set of techniques for obtaining multi-dimensional prediction intervals with finite-sample coverage guarantees. The primary method, \SDI, applies conformal methods to determine a parameter $\beta$, which is applied to scale the sample standard deviation along all dimensions of the data. We then extended \SDI{} to provide prospective prediction intervals for entire trajectories in MDPs. These intervals can be made semi-conditional by first applying quantile regression at each time step and then applying conformalization to the exceedances (the amounts by which each trajectory exceeds the quantile regression intervals at each time step). This results in the \SDT{} method, for which we also provide a proof of correctness. Finally, we demonstrated the method on two application problems: managing tamarisk invasions in river networks and predicting the course of simple Starcraft battles. The experiments in these domains show that these methods perform very well. 

To address safety-critical problems, we developed the \SDTCI{} method, which seeks to ensure that in fraction $1-\delta$ of repeated applications of conformal prediction, the resulting prediction interval achieves coverage of $1-\delta$. This modification was important to achieve good experimental coverage on the Starcraft problem.

Finally, we explored an alternative to trajectory-wise prediction intervals that gives a conformal bound on the total amount by which the trajectory will exceed the quantile regression bounds. We found that this produces reasonably faithful depictions of future behavior for $\delta=0.2$ and $\delta=0.1$ on the Tamarisk problem, and $\delta = 0.2, 0.1$, and $0.05$ on the Starcraft task. For the more stringent guarantees, the total exceedance bounds grow extremely large (especially for Tamarisk). In those cases, the quantile regression bounds are very far from faithfully characterizing the future behavior of the MDP policy.

\subsection{Limitations and Extensions}

The main limitation of the method is the need to estimate extreme quantiles during the conformalization process. Our experiments suggest that at least 500 trajectories are required to obtain reasonably tight estimates even when employing an upper confidence bound to estimate the conformal quantiles. A second limitation is that, unlike the quantile regression bounds, the conformalization adjustments are unconditional, but they may be misinterpreted by users as conditional. We encourage practitioners to inspect the training trajectories that violate the trajectory-wise bounds to verify that the violations do not cluster in particular regions of the input space. If such precautions are taken, the intuitive interpretation as conditional bounds will be very reasonable. We gave examples of this analysis in both the Tamarisk task (where no clusters were found) and in the Starcraft task (where a large cluster of violations was detected). 

The \SDT{} algorithm provides trajectory-wise guarantees for a scalar behavior value $b_t$. In many applications, it would be nice to obtain simultaneous guarantees on multiple behavior variables. This can be easily accomplished by fitting quantile regression functions and computing exceedances for each behavior variable. Then the max exceedance, $c$, should be the maximum of all exceedances of all behavior variables along the trajectory.

In our experiments, we employed quantile random forests to make independent predictions at each time step. The resulting quantile prediction intervals can vary substantially from one time step to the next, and this can lead to very ragged prediction intervals. In problems such as Tamarisk, we know that each trajectory is monotonically decreasing over time, so we would expect the lower bounds to exhibit the same behavior (although this is not a necessary property). Methods for functional quantile regression \citep[e.g.,][]{Beyaztas2021} might provide smoother quantile intervals by simultaneously predicting quantiles along the entire trajectory. 

In both \SDI{} and \SDT{}, we rescale the multiple dimensions and time steps using the sample standard deviation, but many other choices of scaling techniques are possible. For example, \cite{Diquigiovanni2021} employ an initial unscaled conformal method to define a scaling function, and one could also employ the inter-quartile range. Both of these are more robust to outliers than the sample standard deviation.

Finally, in safety-critical problems, it would be helpful if the AI agent could alert the decision maker when the prediction interval is likely to be violated. We saw in Figure~\ref{fig:tamarisk-trajectories}(lower) a case in which multiple attempts to eradicate tamarisk plants failed. This ``unlucky'' sequence of failures causes the agent to violate the prediction interval at time 6. An interesting direction for future research would be to fit survival models that could predict (e.g., by time step 3) that the prediction interval was likely to be violated. This would provide the decision maker with advanced warning, which could allow them to mitigate the problem (e.g., by increasing the available budget for eradication). \cite{Candes2021} show how to provide conformal guarantees for survival analysis.

\section{Acknowledgements} 
This material is based upon work supported by the Defense Advanced Research Projects Agency (DARPA) under Contract No. HR001119C0112. Any opinions, findings and conclusions or recommendations expressed in this material are those of the authors and do not necessarily reflect the views of the DARPA. The authors thank Majid Alkaee Taleghan for help with the Tamarisk model. The authors also thank Kiri Wagstaff, Si Liu, and anonymous UAI 2021 reviewers for comments on earlier drafts.

\vskip 0.2in
\bibliography{conformal-pi}

\clearpage
\appendix

\section{Comparison of Strict and Confidence Interval Quantile Estimation}
\label{ap1}

\begin{figure*}
    \centering
    \includegraphics[width=2.5in]{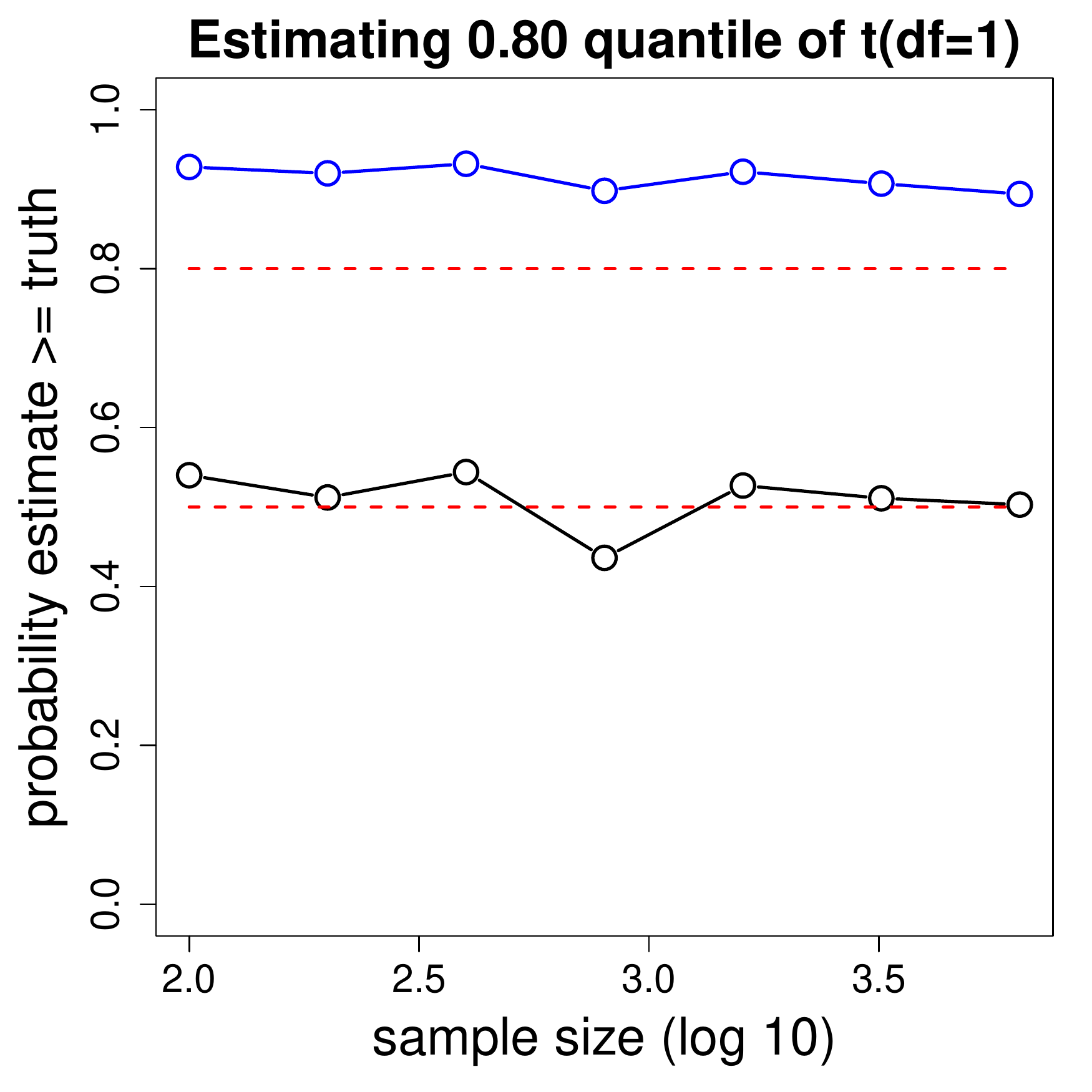}
    \includegraphics[width=2.5in]{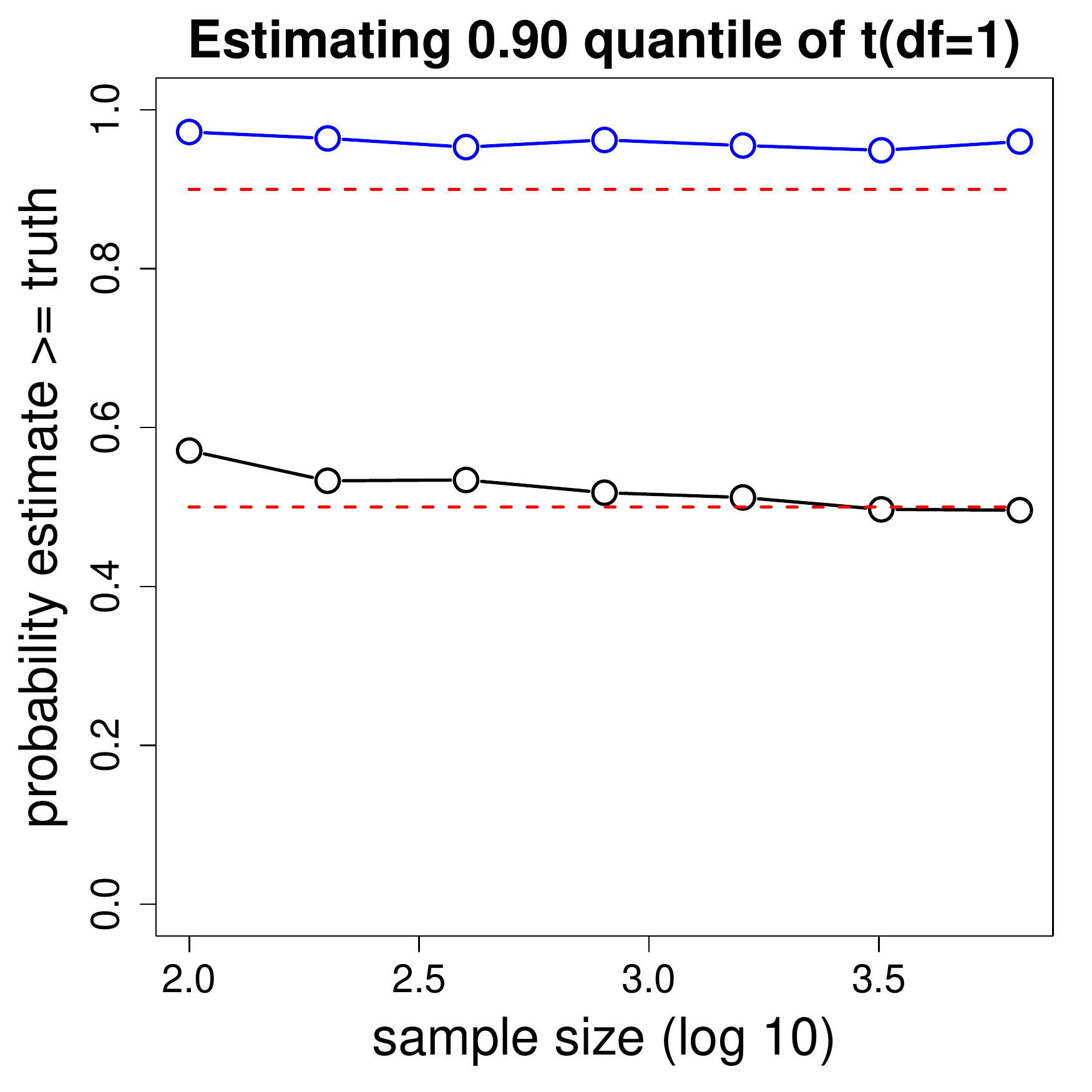}
    \includegraphics[width=2.5in]{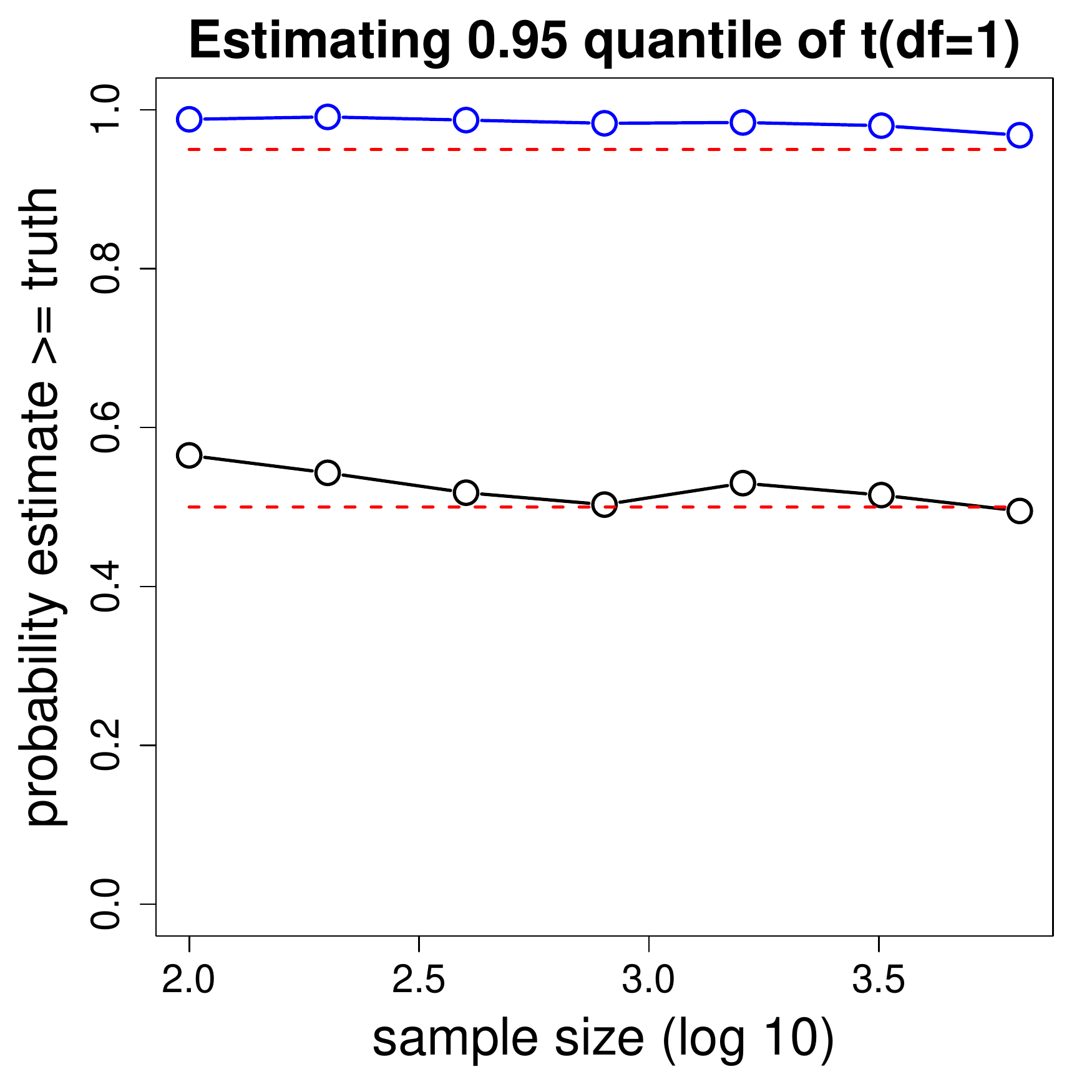}
    \includegraphics[width=2.5in]{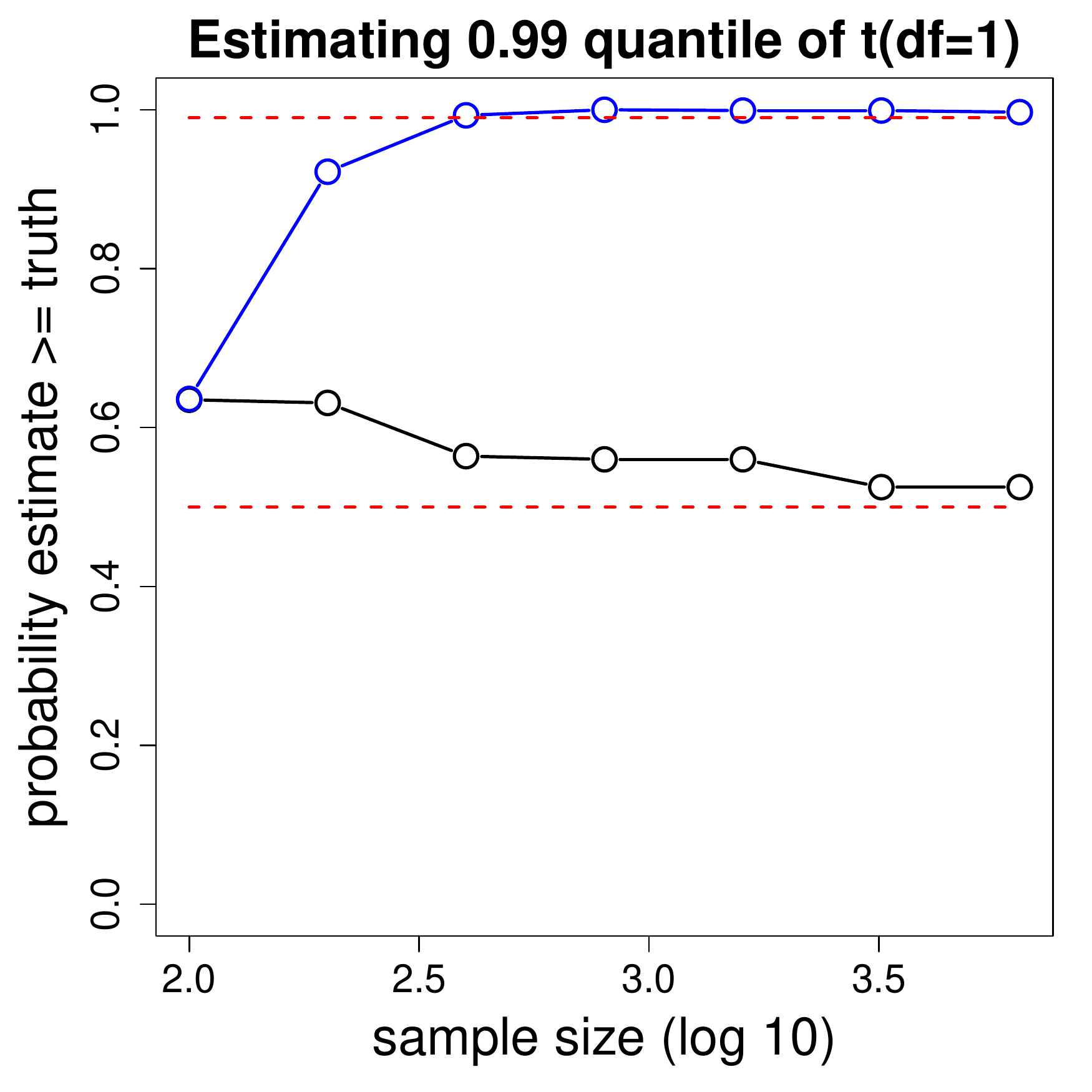}
    \caption{Fraction of 1000 trials in which the standard conformal method (black) and the CI method (blue) match or exceed the true $1-\delta$ quantile of the $t$ distribution with 1 degree of freedom. Lower red dashed line is at 0.5; upper red dashed line is at $1-\delta$.}
    \label{fig:t-success}
\end{figure*}

Our decision to introduce the Confidence Interval approach to estimating the $(1-\delta)((n+1)/n)$ quantile was based on the following simple study. For $\delta \in \{0.2, 0.1, 0.05, 0.01\}$ and $n\in \{200, 400, 800, 1600, 3200, 6400\}$, we generated data points from a Student $t$ distribution with 1 degree of freedom and then compared the theoretical $1-\delta$ quantile of the $t$ distribution to the conformal estimate in 1000 trials. Figure~\ref{fig:t-success} plots the fraction of trials in which the estimated quantile matched or exceeded the true value. The black curves plot the estimates from the standard conformal method ($c_{(\lceil (1 - \delta)(n+1)\rceil)}$) and the blue curves plot the estimates of the $1-\delta/2$ upper confidence bound on the $(1-\delta)((n+1)/n)$  quantile using the method of Nyblom [1992]. We know from the theory that the expected value of the conformal estimate should be equal to the true $1-\delta$ quantile. Here, we observe that in approximately half of the trials, the conformal estimate is smaller than the true quantile and in the other half, it is larger. In a real application, we can only apply the estimation method once, and this suggests that roughly half of the time, the estimated prediction interval will be too small.

In contrast, the $1-\delta$ upper confidence bound on the $(1-\delta)((n+1)/n)$ quantile generally matches or exceeds the true $1-\delta$ quantile at least fraction $1-\delta$ of the time. We refer to this informally as a ``double-$\delta$'' behavior. Only when $1-\delta=0.99$ and the same size is 200 or 400 does the upper confidence estimate fail. 

\section{Code and Data Release}
Upon acceptance, we will publish the code and data sufficient to reproduce all of the results in this paper. 
\end{document}